\documentclass{article}

    \PassOptionsToPackage{numbers, compress}{natbib}
 \usepackage[preprint]{neurips_2026}

\usepackage[utf8]{inputenc} 
\usepackage[T1]{fontenc}    
\usepackage{hyperref}       
\usepackage{url}            
\usepackage{booktabs}       
\usepackage{amsfonts}       
\usepackage{nicefrac}       
\usepackage{microtype}      
\usepackage{xcolor}         

\usepackage{wrapfig}
\usepackage{subcaption}
\usepackage{graphicx}
\usepackage{float}
\usepackage[most]{tcolorbox}
\usepackage{xcolor}
\usepackage[capitalize,noabbrev]{cleveref}
\usepackage{enumitem}
\usepackage[most]{tcolorbox}

\newtcolorbox{runningexample}[1][Running Example]{
  enhanced, breakable,
  colback=white,
  frame hidden,
  borderline west={2pt}{0pt}{black},
  sharp corners,
  fonttitle=\bfseries,
  coltitle=black,
  colbacktitle=white,
  title=#1,
  attach boxed title to top left={xshift=10pt, yshift=0pt},
  boxed title style={
    colback=white, colframe=white,
    boxrule=0pt, sharp corners,
    left=0pt, right=0pt, top=0pt, bottom=0pt,
  },
  top=10pt,
  left=10pt, right=4pt, bottom=4pt,
}

\newcommand{\todo}[1]{}
\renewcommand{\todo}[1]{{\color{red} TODO: {#1}}}
\newcommand{\bench}{\textsc{CL-Bench}}

\makeatletter
\renewcommand\paragraph{\@startsection{paragraph}{4}{\z@}%
  {0pt}{-1em}{\normalfont\normalsize\bfseries}}
\makeatother

\title{\textsc{Continual Learning Bench}: Evaluating Frontier AI Systems in Real-World Stateful Environments}

\author{%
  Parth Asawa\thanks{Correspondence to \texttt{pgasawa@berkeley.edu}. Website: \url{https://www.continual-learning-bench.com}} \\
  UC Berkeley \\
  \And
  Christopher M. Glaze \\
  Snorkel AI \\
  \And
  Gabriel Orlanski \\
  University of Wisconsin-Madison \\
  \And
  Ramya Ramakrishnan \\
  Snorkel AI \\
  \And
  Benji Xu \\
  UC Berkeley \\
  \AND
  Asim Biswal \\
  UC Berkeley \\
  \And
  Vincent Sunn Chen \\
  Snorkel AI \\
  \And
  Frederic Sala \\
  University of Wisconsin-Madison, Snorkel AI \\
  \And
  Matei Zaharia \\
  UC Berkeley \\
  \And
  Joseph E. Gonzalez \\
  UC Berkeley \\
}

\begin{document}

\maketitle

\begin{abstract}
Continual learning, the ability of AI systems to improve through sequential experience, has attracted substantial interest, but no high-quality benchmark exists to evaluate it. We introduce Continual Learning Bench (\bench{}), the first difficult, expert-validated benchmark designed to measure whether LLM-based systems genuinely improve with experience. \bench{} spans six diverse domains (software engineering, signal processing, disease outbreak forecasting, database querying, strategic game-playing, and demand forecasting), each validated by domain experts and designed so that tasks share a learnable latent structure (codebase layout, disease outbreak dynamics, opponent strategies) that a stateful system can discover online but a stateless one cannot. We evaluate frontier models across several agent architectures, from naive in-context learning (ICL) to dedicated memory systems, introducing a gain metric to isolate learning from prior capabilities. We find that these systems leave headroom for improved continual learning: agents frequently overfit to immediate observations or fail to reuse knowledge across instances, and dedicated memory systems do not fix this---in fact, naive ICL outperforms systems dedicated to memory management. \bench{} is the first benchmark to evaluate continual learning across diverse real-world domains with expert-validated tasks and isolate online learning from underlying model capability, showing a need for better continual learning systems.
\end{abstract}

\section{Introduction}
\label{introduction}

Building LLM systems that improve through sequential experience (\emph{continual learning}) has attracted substantial interest from researchers and practitioners alike. Recent work focuses on developing memory-based and adaptive AI systems intended to operate over long time horizons: software engineering agents that become more effective within a codebase over weeks of interaction, data science agents that learn from repeated interaction with the same datasets, and decision-support agents that refine predictions using ongoing feedback. These systems commonly incorporate memory retrieval modules~\citep{packer2024memgptllmsoperatingsystems, mem0, ace}, context compaction methods ~\citep{eyuboglu2025cartridgeslightweightgeneralpurposelong}, and test-time training objectives and architectures ~\citep{sun2025learninglearntesttime, shenfeld2026selfdistillationenablescontinuallearning, hubotter2026reinforcementlearningselfdistillation, tandon2025endtoendtesttimetraininglong, zweiger2025selfadaptinglanguagemodels, lin2025continuallearningsparsememory}.

Yet existing evaluation protocols only partially capture this form of continual learning. Memory and long-context systems are evaluated on recall or question-answering over prior context \citep{maharana2024evaluatinglongtermconversationalmemory, adams2024longhealth, packer2024memgptllmsoperatingsystems, mem0}; compaction methods on whether they preserve full-context behavior while reducing memory or compute \citep{eyuboglu2025cartridgeslightweightgeneralpurposelong, ace}; and test-time training or self-adaptation methods on language-modeling loss, knowledge insertion, or new-task accuracy when provided with training and test sets \citep{sun2025learninglearntesttime, tandon2025endtoendtesttimetraininglong, shenfeld2026selfdistillationenablescontinuallearning, hubotter2026reinforcementlearningselfdistillation, zweiger2025selfadaptinglanguagemodels, lin2025continuallearningsparsememory, asawa2026sievesampleefficientparametriclearning}. These proxies do not directly test whether a system improves online by learning environment-specific latent structure across related tasks.

We introduce \textbf{Continual Learning Bench (\bench{})}, a benchmark for measuring whether AI systems improve through sequential experience, based on expert-validated tasks that are difficult for current models. \bench{} spans six domains (software engineering, signal processing, disease outbreak forecasting, database querying, strategic game-playing, and demand forecasting). Unlike prior benchmarks \citep{maharana2024evaluatinglongtermconversationalmemory, adams2024longhealth, zhong2026skilllearnbenchbenchmarkingcontinuallearning, zheng2025lifelongagentbenchevaluatingllmagents}, \bench{} tasks are developed to contain latent structure that is realistic but not known a priori, minimizing the confound of performance reflecting pretrained knowledge. Agents are challenged to learn this structure and exploit it online. The latent structure can also undergo concept drift, requiring systems to adapt online rather than rely on static capability.

\begin{figure*}[t]
\centering
\includegraphics[width=\linewidth]{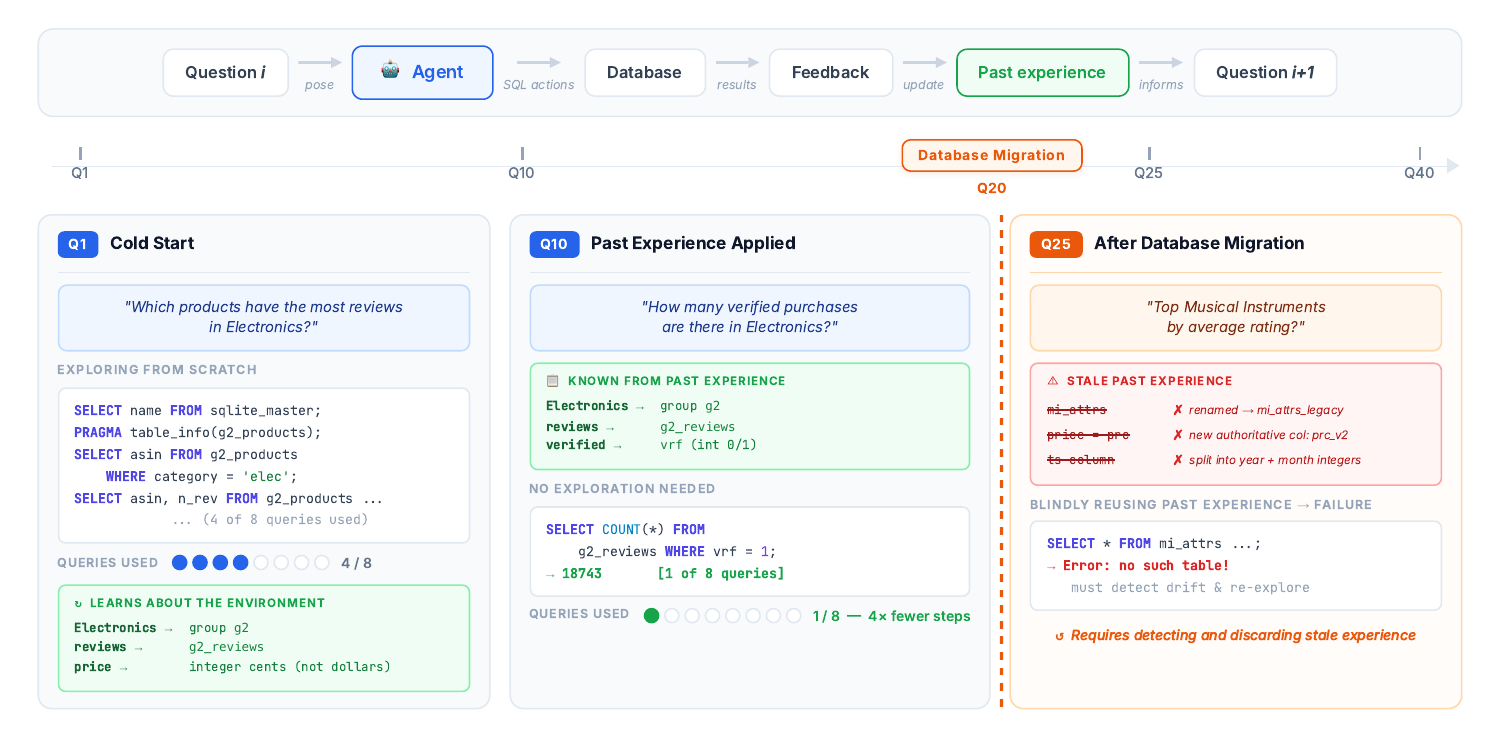}
\caption{%
  \textbf{The \bench{} evaluation framework, illustrated via a simplified Database Exploration task.}
  In each \bench{} task, an agent completes a sequence of instances in a shared environment, accumulating past experience and feedback to improve over time.
  Tasks contain exploitable latent structure not known a priori---here, obfuscated schema conventions---which a learning agent needs to discover to improve reward (e.g. $4\times$ less queries by Q10). Tasks may also feature \emph{concept drift}: a database migration at Q20 invalidates parts of past experience, testing whether agents can adapt rather than blindly memorize.
}
\label{fig:intro-db-example}
\vspace*{-2.5em}
\end{figure*}

\cref{fig:intro-db-example} illustrates these ideas using a Database Exploration task: an agent answers questions about an unfamiliar SQLite database with obfuscated schema conventions and non-standard data formats. A learning agent progressively learns the schema and data---which table group maps to which product category, how prices are encoded---and answers later questions with far fewer queries. Halfway, a database migration introduces concept drift: some past knowledge becomes stale and must be re-learned. We observe a common failure mode where frontier systems are rigid in early, incorrect beliefs, and struggle to update despite feedback on the incorrectness of their answers.

A central challenge in evaluating continual learning is determining what to measure. Each task defines its own reward metric, and intuitively, we want to ensure that continual learning agents are `stateful'. That is, they do better on tasks given prior experience than if running from a fresh start. To measure this, we introduce a gain metric in addition to total reward, which measures the difference between a system's reward conditioned on online experience and what its reward would have been if stateless.

We expect \bench{} to be a continuously improving benchmark, leveraging the open-source community to add more varied and challenging tasks, as well as interesting new systems to test. We define clear criteria in this paper on the qualities of good tasks to make it as easy as possible for contributors to propose additions.

Our contributions are:
\paragraph{\bench{}.} The first difficult, expert-validated continual learning benchmark spanning 6 real-world domains to evaluate how well AI improves from sequential experience, with metrics and tasks designed to isolate learning from base model capability (\cref{sec:cl_bench}, \cref{sec:metric_design}).
\paragraph{Empirical evaluation of the gap in SOTA AI systems.} A large-scale evaluation of frontier models across agent architectures, showing that naive ICL outperforms dedicated continual learning mechanisms, and that reliable online adaptation remains an open problem (\cref{results}).
\section{Related Work}
\label{related_work}

\paragraph{Measuring Continual Learning in Language Models.}
Continual learning has traditionally been evaluated in supervised task sequences using retention, forgetting, and transfer metrics \citep{vandeven2022threetypes}. Recent language-agent benchmarks add interaction and persistence: ARC-AGI-3 measures within-environment skill acquisition \citep{foundation2026arcagi3newchallengefrontier}, LifelongAgentBench and SkillLearnBench structure tasks around predefined skill taxonomies (e.g., named SQL and Bash primitives) to evaluate related task performance and skill reuse \citep{zheng2025lifelongagentbenchevaluatingllmagents, zhong2026skilllearnbenchbenchmarkingcontinuallearning}, and LoCoMo, LongHealth, and MemoryBench test recall fidelity over long conversational histories \citep{maharana2024evaluatinglongtermconversationalmemory, adams2024longhealth, ai2026memorybenchbenchmarkmemorycontinual}. In each case, what the system is expected to learn is either explicitly structured, grounded in general capabilities, or confined to a single stable environment: none require a system to discover latent structure from experience and use this discovery to improve performance on later instances. SWE-Bench-CL is closest, though it is coding-only and not evaluated with frontier systems due to a saturated binary accuracy metric \citep{swebenchcl}. \bench{} instead constructs six expert-validated domains in which the latent structure is hidden, task-specific, and not recoverable from pretraining, so performance improves only when prior experience is correctly exploited.

\paragraph{Agentic Benchmarks.} A number of benchmarks evaluate agentic capabilities such as coding \citep{jimenez2024swebenchlanguagemodelsresolve,swebenchpro,postrainbench}, terminal use \citep{merrill2026terminalbenchbenchmarkingagentshard}, generic computer use \citep{financeagent,osworld,theagentcompany}, and more \citep{zhou2024webarenarealisticwebenvironment,browsecomp}. However, these evaluate agents on isolated tasks and test capabilities learned offline. More recent benchmarks measure performance in \textit{iterative} settings \citep{taubench, slopcodebench}, but similarly don't require exploiting shared latent structure that recurs across instances, so that experience with earlier instances transfers to later ones. Without such a structure, a stronger static model may simply do better, preventing the isolation of continual learning. Our work explicitly designs tasks in which shared latent structure exists and is not a general capability, rewarding systems that improve through online experience.
\section{Continual Learning Bench}
\label{sec:cl_bench}

The key to any effective continual learning benchmark is that the system undergoing evaluation is expected to update as a result of experience or feedback from the environment. As a result, tasks and their corresponding metrics must be designed in such a way that they reward systems which are able to effectively leverage prior online task experience to improve.

\textbf{Terminology}: A \textit{task} in \bench{} is a sequence of problem instances. An \textit{instance} is a single problem within a task (e.g., one question to answer about the database) and the level at which reward is defined. A \textit{step} is a single agent action within an instance (e.g., one SQL query). A \textit{system} is the end-to-end unit under evaluation---any combination of model and learning mechanism is admissible; \bench{} does not prescribe a specific adaptation mechanism. Each task can also define one or more \textit{variants} that modify the environment to introduce concept drift and require systems to further adapt. A \textit{schedule} is the full ordered sequence of instances a system processes from start to finish, spanning multiple variants in a specified order.

\subsection{Task Admission Criteria}
\label{task_formulation}

Any task in an effective continual learning benchmark must satisfy the following criteria. Jointly, they define when measured improvement reflects online learning rather than static model capability.

\paragraph{Headroom.} Initial performance for any system should be well below the achievable maximum. This means the latent structure to be learned must be task-specific and not recoverable from general offline training alone; otherwise, a stronger base model and a learning system are indistinguishable.

\paragraph{Shared latent structure.} There must exist discoverable structure (such as codebase structure, schema conventions, opponent strategies) that is \emph{shared} across instances and that a system can exploit for performance improvement. The relationship between instances creates the learning opportunity: a system that identifies recurring patterns can exploit them in future instances. The structure is not communicated explicitly; it must be inferable from experience.
    
\paragraph{Learning mechanism.} Earlier instances in the schedule must produce observations that are informative for later ones. The environment must provide a feedback loop that a learning system can exploit (e.g., test failures, error messages, intermediate predictions) so that accumulated experience translates into actionable knowledge.

\begin{tcolorbox}[colback=blue!5!white,colframe=blue!75!black,title=Running Example: Database Exploration]
An agent answers 40 natural-language questions about an unknown SQLite database of Amazon product reviews with intentionally obfuscated schema (e.g., table \texttt{g2\_reviews}, column \texttt{vrf} for verified status, prices as integer cents for Electronics but float dollars for other groups). The agent must issue SQL queries to answer each question within a budget of 15 exploratory queries, and \textit{efficiency}---fraction of the budget saved---is the reward.\\

A learning agent discovers schema conventions and data idiosyncrasies early and reuses this logic to answer later questions with far fewer queries. The database also undergoes a migration halfway through, renames tables and reformatting columns, requiring the agent to detect and discard stale past experience rather than blindly reusing it.
\end{tcolorbox}

Database Exploration satisfies all these criteria. Significant headroom exists due to specifically defined schema obfuscation and data idiosyncrasies that are realistic but not known a priori. The shared latent structure is exploitable by a learning system, and the iteration loop provides a way to infer schema and data-value information. The drift additionally tests whether the system can adapt.

Importantly, these criteria jointly preclude constructing \bench{} tasks by chaining instances from existing benchmarks. Standard benchmark instances are designed to be diverse and independent, and test offline training, meaning a sufficiently capable model could perform well without any online learning. \bench{} tasks require improvements that are only achievable by learning from the specific sequential experience of the run. As a result, each task is designed from the ground up: benchmark contributors identify a domain with exploitable latent structure, construct a realistic sequence of instances, verify that the structure is not a general capability, and validate the design with domain experts (see \cref{sec:task_validation}).

\subsection{Task Validation}
\label{sec:task_validation}

\begin{figure}[h]
\begin{center}
\includegraphics[width=0.88\linewidth]{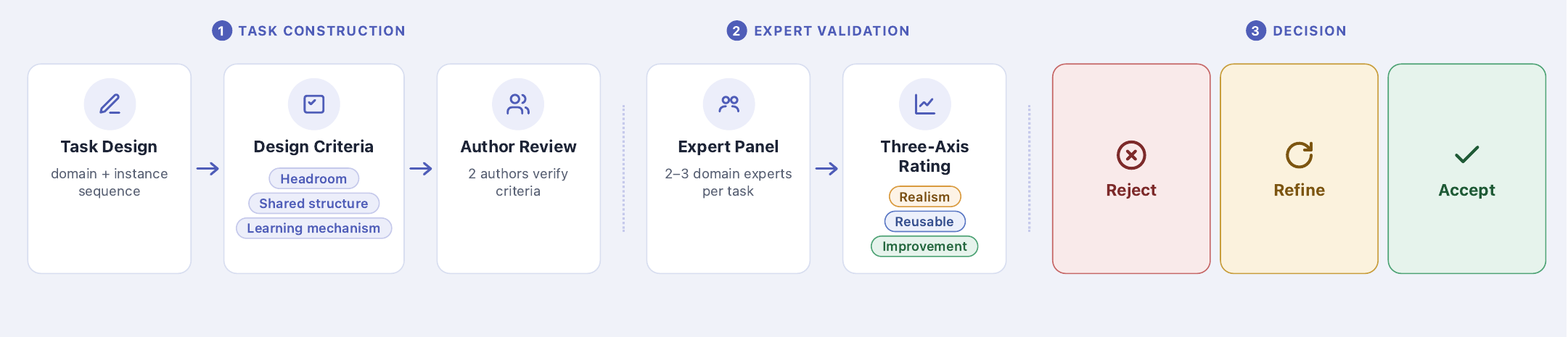}
\end{center}
\caption{\bench{} task construction and human validation pipeline. Each task is constructed against the 3 design criteria and reviewed by at least two authors, then independently reviewed by 2–3 domain experts on three axes (realism, reusable knowledge, and learning improvement) with tasks accepted into the benchmark only after confirming validity on all axes.}
\label{fig:task_construction_and_validation}
\vskip -0.1in
\end{figure}

After a task is proposed and meets the above criteria, we take additional steps to validate and ensure it is suitable for testing continual learning. First, each task undergoes a manual review by two authors to ensure that the designed task satisfies the criteria desired for testing continual learning systems.

Then, to ensure that domain-specific tasks reflect realistic workflows and present real learning challenges, we engage with domain experts through a structured validation process. Two to three domain experts reviewed each task specification against the criteria below.

\noindent\textbf{Realism.} Does the task reflect a realistic real-world workflow, with inputs, steps, and feedback that a domain practitioner would genuinely encounter? Tasks should avoid artificial puzzle structure, over-simplified setups, or hints that would not exist in practice.

\noindent\textbf{Reusable Knowledge.} Does the task require knowledge accumulation across steps? Specifically, do patterns, context, or strategies discovered during early steps transfer to later steps, such that a system with memory would outperform one without?

\noindent\textbf{Learning Improvement.} Would performance measurably improve over time? Tasks should avoid scenarios in which the improvement signal is too subtle or noisy to be learned over the task horizon.

Experts also answered task-specific questions and provided open-ended feedback used to iteratively refine each task. On average, two to three experts each spent 2.14 hours per cycle on task validation, with times ranging from 1 to 4.5 hours.

\paragraph{Tasks are learnable but unsolved.} While tasks present learning opportunities, the results in \cref{sec:core_results} empirically verify that current frontier systems leave headroom unrealized. 
Though there is no clearly defined optimal learning algorithm for many of these tasks, it is clear that systems dedicated to memory management are worse than the simplest in-context learning baseline. Even ICL indicates consistent learning deficits, over-relying on recent task instances and under-weighting less recent task variants that are relevant to inference and performance; see \cref{sec:core_results} for more.

\subsection{Benchmark Construction}
\label{sec:construction}

\begin{figure}[h]
    \centering
    \includegraphics[width=.9\linewidth]{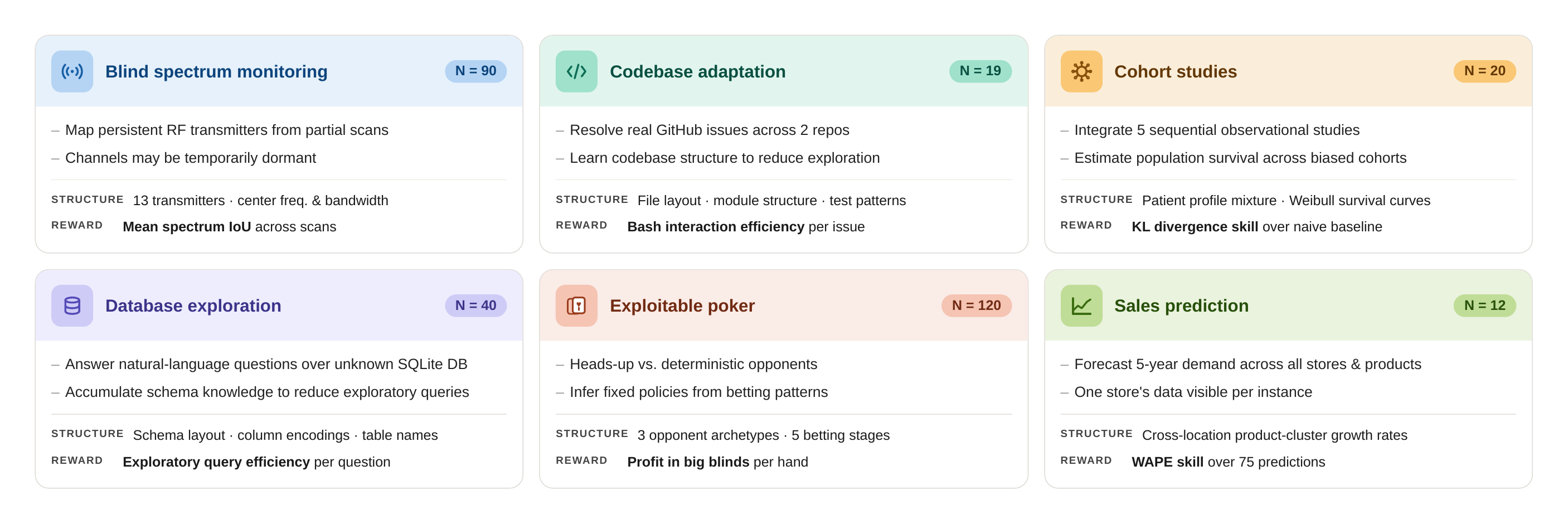}
    \caption{\textbf{\bench{} task overview.} $N$ is the number of instances per rollout (with each instance consisting of up to 40 steps depending on the task). Latent structure is what a learning system can discover and exploit across instances; reward is the axis along which performance is tracked.}
    \label{fig:benchmark_tasks}
    \vskip -0.15in
\end{figure}

\cref{fig:benchmark_tasks} summarizes the task sets introduced in \bench{}. Full descriptions, instance examples, variant definitions, and further curation details are provided in Appendix~\ref{appendix_task_details}.

The tasks span six domains---software engineering, database analytics, epidemiological forecasting, RF spectrum monitoring, sales forecasting, and strategic gaming---with reward types ranging from step efficiency and prediction accuracy to signal quality and financial profit. Interaction styles are similarly varied: two tasks run agents in persistent terminal environments with full shell access (Codebase Adaptation, Sales Prediction), one exposes a structured tool API (Cohort Studies), and all tasks require structured outputs for reproducible automated scoring. This breadth ensures that \bench{} stresses continual learning across realistic workflows rather than a single task format.

\section{Metric Design}
\label{sec:metric_design}
We choose two axes to analyze performance: reward and gain, which we describe below.

\subsection{Reward}
\label{sec:reward}

A \textbf{learning system} $\mathcal{S}$ processes a schedule of $N$ instances $x_1, \ldots, x_N$ sequentially; at each instance $t$ it receives $x_t$, produces a response, and observes reward $r_t \in \mathcal{R}$. \textbf{Stateless} ($sl$) systems condition only on the current instance, $\pi_t = \pi(x_t)$, independent of history $\mathcal{H}_{t-1} = \{(x_i, r_i)\}_{i=1}^{t-1}$; by construction, they cannot improve across instances within a rollout. \textbf{Stateful} ($sf$) systems condition on the full history, $\pi_t = \pi(x_t, \mathcal{H}_{t-1})$, where state may be maintained explicitly (e.g., retrieved memory) or implicitly (e.g., accumulated context).

A higher reward $r_t$ reflects a better outcome for an instance and is directly tied to the task's objective (e.g., exploration efficiency, minimization of prediction error). Tasks differ in their natural reward scale as a result, but all satisfy these properties by construction. Specific details for each task's reward definitions are provided in Appendix \ref{appendix_task_details}. The per-task reward aggregate is the cumulative reward across all $N$ instances in the schedule for a stateful system ($sf$): $r^{sf} = \sum_{t=1}^{N} r^{sf}_t$.

When instance difficulty is homogeneous, a rising reward curve is a reliable signal of learning. In practice, however, instance difficulty varies, and different systems have different baseline performance. A generically strong system may produce high reward with no learning at all while a weaker system that exploits prior experience may show lower absolute reward but actually improves over time. To distinguish between these, we introduce the \textbf{gain} metric.

\subsection{Gain}
\label{sec:gain}

\begin{figure}[H]
    \centering
    \includegraphics[width=0.75\linewidth]{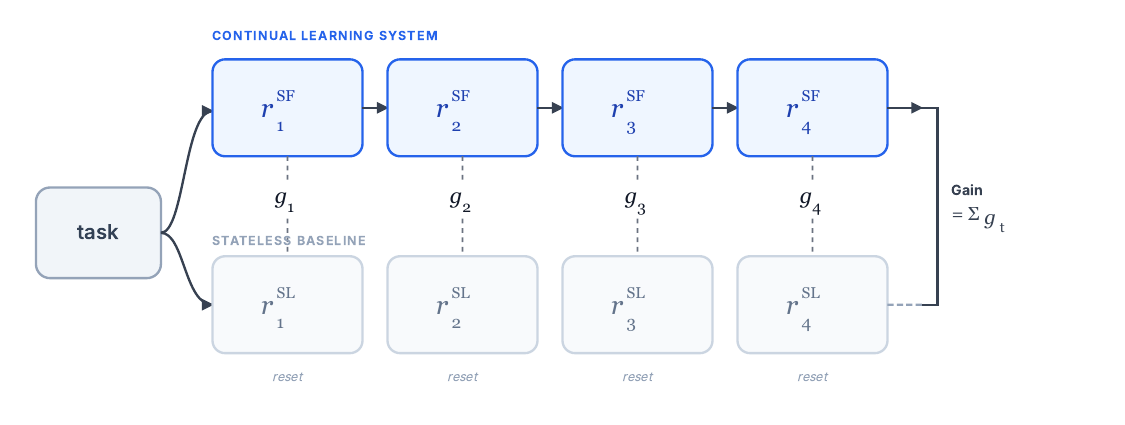}
    \caption{\textbf{Overview of the gain metric.} We compute the difference between a system's stateful performance on task instances vs. its performance if it had seen each instance independently. This allows us to isolate the performance improvement due to learning from prior experience.}
    \label{fig:gain_diagram}
    \vskip -0.2in
\end{figure}

For each instance $t$, we define gain as: $g_t = r^{sf}_t - r^{sl}_t$ where $r^{sf}_t$ is the reward achieved by the stateful system and $r^{sl}_t$ is the reward achieved by the same system run in stateless mode on the same instance. Whatever makes instance $t$ intrinsically easy or hard for a given system appears equally in both terms. The difference isolates only the contribution of the accumulated state to performance. The per-task gain aggregate is the cumulative gain $g = \sum_{t=1}^{N} g_t$.

\subsection{Cross-Task Normalization}
\label{sec:cross_task_norm}

A good continual learning benchmark should be task-agnostic, and thus, to provide a single ranking for systems across our tasks, we must aggregate metrics across tasks, which requires normalizing heterogeneous reward scales. To do so, each task specifies a maximum possible average reward obtainable $r_{max}$. Each task's $r_{max}$ is defined in Appendix~\ref{appendix_task_details} alongside the scoring formula. Normalized gain is then: $\widehat{g} = \frac{\bar{r}^{sf} - \bar{r}^{sl}}{r_{max} - \bar{r}^{sl}}$

The denominator $(r_{max} - \bar{r}^{sl})$ is the system's own learning headroom—the maximum gain available given its stateless baseline performance. Dividing by headroom prevents tasks where the stateless baseline is already near $r_{max}$ from contributing a negligible signal regardless of actual learning, and ensures every task is on a 'fraction of available headroom captured' scale. This normalized gain measure has precedent in education \citep{hake1998interactiveengagement} and has been adapted for continual RL benchmarking \citep{wolczyk2021continualworldroboticbenchmark}.

For normalized reward, we use a fixed external reference $r_{external}$, the stateless ICL reward of a frontier model (GPT-5.4 ICL), rather than each system's own stateless performance: $\widehat{r} = \frac{\bar{r}^{sf} - r_{external}}{r_{max} - r_{external}}$

The external reference makes reward scores submission-independent: a system's normalized reward does not change as new systems are added to the benchmark.
\section{Evaluating Frontier Systems on \bench{}}
\label{results}

We evaluate frontier language models with agent architectures designed to enable continual learning on our benchmark. All systems are averaged over 5 rollouts per task in potentially different instance orders to reduce spurious ordering effects. Further details on noise reduction are in Appendix ~\ref{sec:noise}.

\label{sec:experimental_setup}

\paragraph{Models.} The backbone of all systems we evaluate is an LLM. We test frontier models including: Claude Opus 4.7~\citep{opus47} and Sonnet 4.6~\citep{sonnet46}, Gemini 3.1 Pro~\citep{gemini31} and Gemini 3 Flash~\citep{gemini3flash}, and GPT 5.4~\citep{gpt54} in various continual learning systems (below). For all models, we use default reasoning settings. Models are queried via native provider APIs, with prompt caching enabled where possible.

\paragraph{Continual Learning Systems.} We evaluate systems spanning the major memory paradigms available to LLM-based agents. Full-context ICL preserves the entire conversation history without truncation unless context windows are exceeded, providing a strong context-based baseline. ICL Notepad gives the agent a persistent scratchpad: at each turn it updates structured notes injected at the start of the next step, placing the burden of memory curation on the model itself. Mem0~\citep{mem0} instead automates this by semantically extracting memories each step and retrieving the top-$k$ most relevant ones for each new step (we set $k=10$). ACE~\citep{ace} maintains a playbook of memories that is updated after each instance, enabling persistence of prior experience. We additionally evaluate Claude Code~\citep{anthropic_claude_code_2025} and Codex~\citep{openai_codex_2025}, agentic harnesses invoked in headless mode with a single persistent conversation per task following the TerminalBench~\citep{merrill2026terminalbenchbenchmarkingagentshard} implementation. Both apply automatic context compaction as conversations grow. We further detail our system-model pairs in Appendix~\ref{sec:sys_selection}.

\subsection{Results}
\label{sec:core_results}

\begin{table}[b]
\caption{\textbf{ICL is a strong baseline continual learning system.} Normalized reward is averaged across the six benchmark tasks after subtracting the stateless GPT-5.4 baseline and scaling by the task-specific maximum reward. Normalized gain instead subtracts each system's own stateless baseline. Cost reports the mean total rollout cost across tasks in USD. 95\% CIs from per-task rollout variance are provided for all systems except for Codex due to resource limitations.}
\label{tab:overall-results}
\begin{center}
\begin{small}
\begin{sc}
\resizebox{\linewidth}{!}{
\begin{tabular}{c|llccc}
\toprule
Rank & System & Model & Norm. reward (\%) & Norm. gain (\%) & Cost (\$) \\
\midrule
1 & ICL & Claude Sonnet 4.6 & \textbf{22.3} $\pm$ 4.1 & \textbf{25.4} $\pm$ 3.6 & 30.4 \\
2 & ICL & GPT-5.4 & 20.1 $\pm$ 9.1 & 20.1 $\pm$ 9.1 & 18.4 \\
3 & Claude Code & Sonnet 4.6 & 19.0 $\pm$ 7.1 & 23.9 $\pm$ 5.7 & 38.6 \\
4 & Mem0 & GPT-5.4 & 15.1 $\pm$ 6.4 & 20.2 $\pm$ 5.9 & 18.3 \\
5 & ICL & Claude Opus 4.7 & 10.2 $\pm$ 4.4 & 19.5 $\pm$ 4.1 & 49.6 \\
6 & ICL Notepad & GPT-5.4 & 8.0 $\pm$ 3.3 & 7.8 $\pm$ 3.0 & 14.3 \\
7 & ICL & Gemini 3 Flash & 8.0 $\pm$ 4.7 & 16.4 $\pm$ 3.8 & \textbf{7.6} \\
8 & Codex & GPT-5.4 & 6.6 & 14.6 & 27.2 \\
9 & ACE & GPT-5.4 & 4.6 $\pm$ 2.7 & 8.6 $\pm$ 2.5 & 62.8 \\
10 & ICL Notepad & Claude Sonnet 4.6 & 3.5 $\pm$ 5.7 & 18.2 $\pm$ 3.4 & 31.5 \\
11 & ICL Notepad & Gemini 3.1 Pro & -0.2 $\pm$ 3.4 & 9.4 $\pm$ 3.0 & 13.3 \\
12 & ICL & Gemini 3.1 Pro & -5.6 $\pm$ 4.1 & 6.2 $\pm$ 3.8 & 15.2 \\
\bottomrule
\end{tabular}
}
\end{sc}
\end{small}
\end{center}
\end{table}

\paragraph{ICL serves as a strong baseline.}~\cref{tab:overall-results} shows that full-context ICL with Claude Sonnet 4.6 achieves the highest aggregate normalized reward (22.3\%) and gain (25.4\%) of any evaluated system. More broadly, ICL-based systems occupy three of the top five positions in terms of gain. In contrast to this, ICL Notepad, using the same Sonnet 4.6 model, ranks sixth overall in gain (18.2\%) with a substantially lower absolute reward (3.5\%), indicating that the choice of learning medium matters as much as the underlying model, and that the gain and reward metrics are both important in capturing both absolute performance and the learning from the system. Other dedicated agent memory systems do poorly relative to their cost: ACE ranks tenth by gain (8.6\%) while incurring the highest cost of any system (\$62.8 per full run), and ICL Notepad with GPT-5.4 ranks last among systems using the same model. The simplest mechanism of preserving context without compression remains effective.

\begin{figure*}[t]
\begin{center}
\centerline{\includegraphics[width=0.88\linewidth]{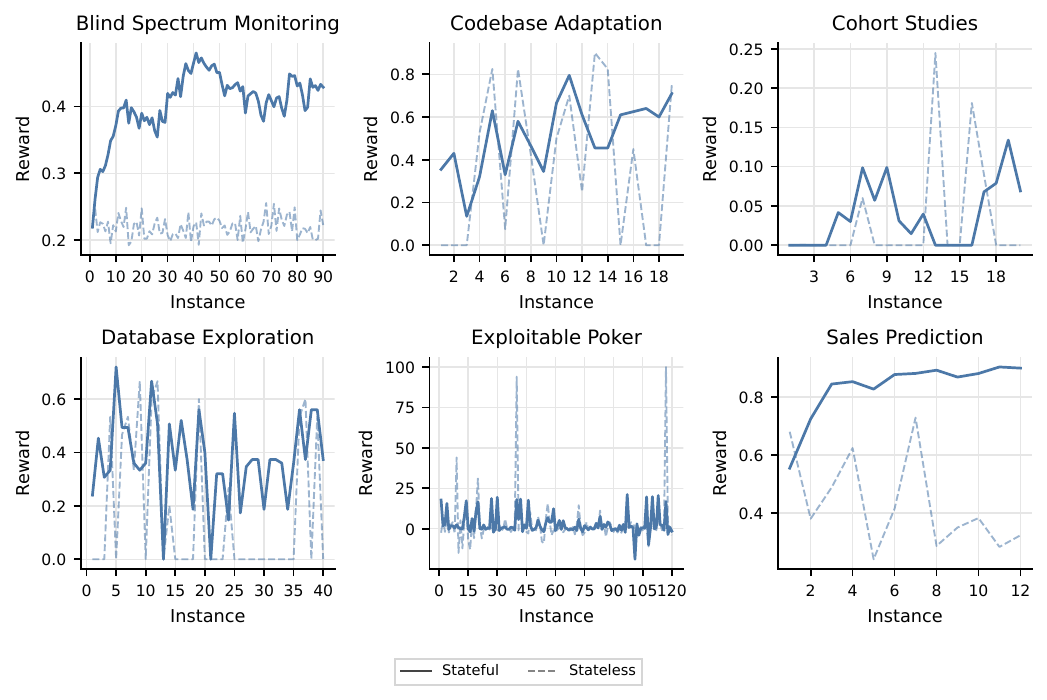}}
\caption{\textbf{Learning curves across all six tasks for the top-ranked system (ICL + Claude Sonnet 4.6).} Solid lines show mean reward per instance when stateful; dashed lines show the same system run statelessly. The gap between lines isolates the contribution of learning from experience to performance. Sales Prediction and BSM show the largest and earliest-forming gaps; whereas Database Exploration is notable for the frequency with which stateless reward collapses to zero in contrast to stateful; Cohort Studies, Codebase, and Poker show minimal gain throughout with both curves overlapping.}
\label{fig:task-curves-top-system}
\end{center}
\vskip -0.2in
\end{figure*}

\paragraph{Per task learning curves.} \Cref{fig:task-curves-top-system} reveals heterogeneity in learning dynamics across tasks. Sales Prediction and Blind Spectrum Monitoring show the clearest learning: stateful reward in Sales Prediction rises by the third instance and holds while stateless remains volatile; BSM shows steady accumulation as the channel map builds. Database Exploration illustrates a different mechanism—stateless reward frequently collapses to zero while stateful maintains a floor, so gain is visible through baseline failure rather than stateful improvement. Cohort Studies shows the least amount of learning with both curves hovering near zero performance. This reflects challenging cross-study inference (different studies in different variants), with each individual study containing confounding variables that are typical in epidemiology. Experts validated the learnability of the latent structure but no evaluated system currently extracts it, appearing misled by spurious correlations.

\begin{figure*}[t]
\vskip -0.1in
\begin{center}
\centerline{\includegraphics[width=0.88\linewidth]{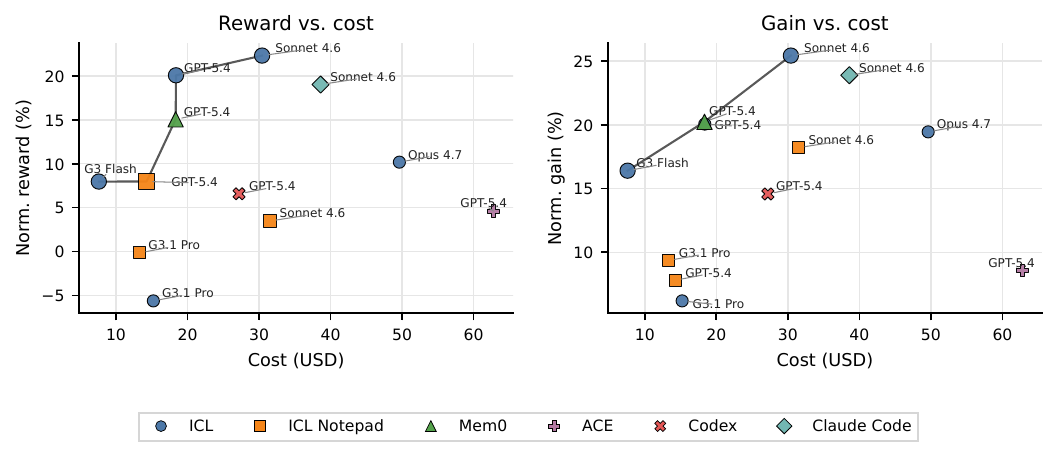}}
\caption{\textbf{Aggregate Pareto views of the public \bench{} leaderboard.} We compare the normalized reward against the rollout cost and the normalized gain against the rollout cost. Black lines connect configurations not dominated by another configuration on the plotted axes. Model names that start with solely G are shorthand for Gemini models.}
\label{fig:pareto-frontiers}
\end{center}
\vskip -0.4in
\end{figure*}

\paragraph{Cost efficiency favors ICL.} The pareto plots in Figure~\ref{fig:pareto-frontiers} show that no dedicated memory system dominates ICL on either cost axis. ICL with Gemini Flash offers the best cost efficiency at \$7.6 per run with 16.4\% gain—comparable to more expensive configurations. Systems that incur high cost through test-time compute (ACE: \$62.8, Claude Code: \$38.6) do not translate that cost into proportionally higher gain. The frontier on the gain-vs-cost plot is largely occupied by ICL variants, with Claude Code providing the one clear exception: it achieves the second-highest gain (23.9\%) at moderate cost (\$38.6) while also delivering strong per-task gains on Sales Prediction (65.1\%) and Database Exploration (43.6\%) that no ICL configuration matches on those tasks individually.

\paragraph{Memory systems show different learning deficits.} We also measure how much learning gain is driven by adapting to new information versus retaining older information across the task schedule. This plasticity-stability trade-off is well documented in in the literature \citep{van_de_Ven_2025, delange2023stabilitygap, delange2022survey}.
We decompose gain into two components using variant boundaries as a natural partition: stability measures gain at the first instance of each new variant, where the system must transfer prior knowledge without in-variant feedback; plasticity measures gain within variants, where in-variant feedback enables adaptation. A system that retains useful structure across variant switches shows positive stability; one that adapts quickly within a variant shows positive plasticity (see Appendix~\ref{appendix:learning_decomposition} for formal derivation).
  
\begin{figure*}[t]
\vskip -0.1in
\begin{center}
\centerline{\includegraphics[width=0.77\linewidth]{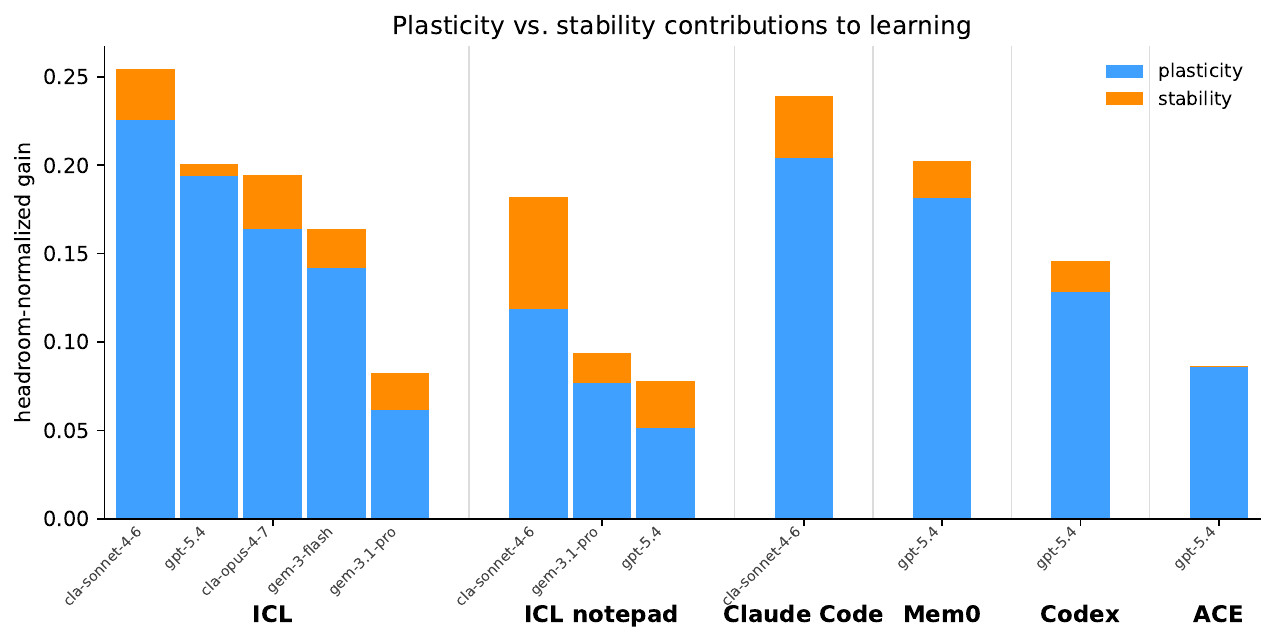}}
\caption{\textbf{Decomposition of normalized gain into stability and plasticity terms.} Normalized gain for each system and agent decomposed into (1) a stability term that measures how effective the agent is at re-using previously acquired information and (2) a plasticity term that measures how effective the agent is at incorporating new information to perform tasks.}
\label{fig:learning-decomposition}
\end{center}
\vskip -0.3in
\end{figure*}

With this decomposition, we indeed see significant variance in how much each component contributes to aggregate learning across agents and systems (\cref{fig:learning-decomposition}), with the ICL Notepad system indicating the most "stable" learning overall with Claude Sonnet 4.6, compiling observations from task outcomes though not adapting as well to new information. ICL and Claude Code indicate the most amount of "plastic" learning, indicating relatively fast adaptation to task variants. Interestingly, some systems indicate virtually no stable learning (ACE and GPT 5.4 enabled with ICL), suggesting that most of the performance gain is those systems is driven by repeated exposure to the same task variants, offset by information loss after switches to new variants.  We provide examples of these in Appendix~\ref{appendix:failure_examples}.

\section{Limitations}
\label{sec:limitations}

\bench{} covers six domains, which, while diverse and expert-validated, do not fully span real-world continual learning scenarios. Task sequences are on the order of tens of instances, which, although they contain many steps, are shorter than the longest deployment horizons practitioners might ultimately care about. However, longer horizons also entail higher costs and runtime, as well as greater validation difficulty, reducing the benchmark's usability. Our initial evaluation focuses on context-based memory paradigms (in-context retention or compaction, retrieval-augmented memory, and structured notepads); we do not evaluate parametric approaches such as test-time training, though we seek to add them via community contributions as researchers develop parametric methods for realistic continual learning settings. Finally, because \bench{} tasks require frontier-level capability to perform non-trivially, failure modes of smaller models may not be surfaced.
\section{Conclusion}
\label{conclusion}

We introduce \bench{}, the first difficult, expert-validated benchmark for measuring whether AI systems genuinely improve through sequential experience. Spanning six real-world domains with verifiable rewards and analysis metrics that isolates learning from underlying model capability, \bench{} enables rigorous comparison of continual learning strategies at frontier scale that is increasingly important as AI agents are deployed in online settings. We expect \bench{} to grow with community contributions in varied and more difficult domains, and define clear criteria for what proposed tasks in a high-quality continual learning benchmark should look like to enable this.

Our evaluation reveals a gap in current systems' ability to continually learn. Naive ICL outperforms dedicated memory architectures on most tasks, and even the best system achieves only 25.4\% normalized gain over its stateless baseline. Accumulated state frequently hurts rather than helps: memory modules introduce spurious generalizations and stale beliefs, while more expensive systems fail to translate cost into performance. These results suggest that continual learning in LLM-based agents remains an open problem, and we hope \bench{} provides the foundation to track progress.

\begin{ack}
We would like to thank Jessy Lin, Braden Hancock, Chris Rytting, Alex Shaw, Ryan Marten, Tyler Griggs, Benedikt Stroebl, Lakshya A Agrawal, Greg Kamradt, Drew Breunig, John Yang, Ofir Press, Abhay Singhal, Dacheng Li, Alan Zhu, K. Tighe, Jason Ding, and Andrew Qin for insightful discussions about this work.

This work is supported by Snorkel AI through the Open Benchmarks Grants program and the Laude Institute via the Laude Slingshots program. Sky Computing Lab is supported by gifts from Accenture, AMD, Anyscale, Cisco, Google, IBM, Intel, Intesa Sanpaolo, Lambda, Microsoft, NVIDIA, Samsung SDS, SAP, and VMware.
\end{ack}

\bibliography{main}
\bibliographystyle{plainnat}


\appendix
\section{Task Details}
\label{appendix_task_details}

This appendix describes each \bench{} task in detail: its continual-learning framing, the information available to the agent each instance, an illustrative prompt–response exchange, the schedule and any concept drift, and how the per-instance reward is computed.

\subsection{Blind Spectrum Monitoring}
\label{appendix:bsm}

\paragraph{Continual learning framing.}
A spectrum monitoring analyst is tasked with maintaining a running estimate of available channels over a 168 MHz band. The analyst receives multiple radio-frequency scans over time (total of 90 for the task) that reveal signals from transmitters occupying the spectrum (so unavailable regions). Importantly, transmitters can be active for long runs and then dormant for long runs (with unpredictable switches in status). The agent must therefore maintain a running map of the \emph{persistent} set of transmitters accounting for this uncertainty, updating and reporting out beliefs about available regions after every scan. 

Here, the latent structure the stateful agent must learn is the full set of available channel center frequencies and bandwidths. Once a channel has been observed, the agent should carry it forward even during periods of silence. A stateless agent that only makes inferences on signals visible in the current scan will systematically miss dormant channels and converge to a lower accuracy than an agent that accumulates evidence across scans.

\paragraph{Schedule and concept drift.}
The default schedule is a 90-scan, three-stage curriculum:

\begin{itemize}[noitemsep]
  \item \textbf{Stage 1} (scans 1--30, \texttt{five\_ch\_wide}): five wideband channels dominate; the narrowband grid is almost entirely dormant.
  \item \textbf{Stage 2} (scans 31--60, \texttt{five\_plus\_four\_mixed}): four narrowband channels begin appearing in guard-band gaps between the wideband channels.
  \item \textbf{Stage 3} (scans 61--90, \texttt{full\_grid\_active}): the full mixed grid of 13 channels becomes active.
\end{itemize}

From the agent's perspective the distribution of detected peaks shifts at each stage boundary, resembling concept drift. The drift is epistemic rather than real: the union of all channel definitions across all three stages is fixed and is what the scorer measures against. What expands is the latent set's \emph{observability}. 

\paragraph{Example interaction.}\leavevmode
\begin{tcolorbox}[colback=gray!5!white, colframe=gray!50!black,
    fonttitle=\bfseries, fontupper=\small\ttfamily, title=BSM: Scan 7 of 90 (Stage 1)]
--- Scan 7/90 ---

Scan metadata: scan\_id: bsm-0007 | timestamp\_utc: 2024-01-07T00:07:00Z
  sensor\_id: SEN-A | integration\_time\_ms: 100 | est.\ noise\_floor: -95 dBm

Detected peaks:
  - peak\_id: p1 | freq: 428.4 MHz | power: -52 dBm | width: 18.0 MHz
  - peak\_id: p2 | freq: 461.2 MHz | power: -48 dBm | width: 22.0 MHz

Band: 340--508 MHz.  Submit your report.
\end{tcolorbox}

\begin{tcolorbox}[colback=blue!4!white, colframe=blue!50!black,
    fonttitle=\bfseries, fontupper=\small\ttfamily, title=BSM: Agent response (after 6 prior scans)]
\{
  "transmitters": [
    \{"center\_freq": 428.4, "bandwidth": 18.0,
     "currently\_active": true,  "estimated\_power": -52.0\},
    \{"center\_freq": 461.2, "bandwidth": 22.0,
     "currently\_active": true,  "estimated\_power": -48.0\},
    \{"center\_freq": 395.0, "bandwidth": 20.0,
     "currently\_active": false, "estimated\_power": -60.0\}
  ]
\}
\end{tcolorbox}

A learning agent reports the third transmitter (center 395 MHz) as \texttt{currently\_active: false} because it appeared in two of the preceding six scans but is absent now---evidence of dormancy rather than non-existence.

\paragraph{Scoring.}
Each scan is scored by the intersection-over-union (IoU) of the agent's reported occupied spectrum and the ground-truth occupied spectrum. IoU equals $|A \cap B| / |A \cup B|$ measured in MHz; it is 1.0 for a perfect report and falls toward 0 as the report diverges from ground truth. The per-run reward is the mean IoU across all 90 scans.

\paragraph{Reference maximum.}
$r_{max} = 1.0$, achieved by a system that correctly reports the full channel set (with accurate center frequencies and bandwidths) at every scan.

\subsection{Codebase Adaptation}
\label{appendix:codebase}

\paragraph{Continual learning framing.}
The agent receives a sequence of 19 real-world Python bugfix issues drawn from two open-source repositories: 9 from \texttt{jazzband/tablib} and 10 from \texttt{jd/tenacity}. Each issue is presented in a fresh Docker container at the issue's base commit; the agent interacts via bash commands and submits a patch, which is evaluated against hidden regression tests. The key learning opportunity is \emph{efficiency}: issues from the same repository often touch overlapping modules, share test invocation patterns, and follow the same contributor conventions. An agent that records where core logic lives, which test commands are reliable, and how maintainers structure fixes needs fewer bash interactions per issue as the sequence progresses---its stateful version outperforms the stateless one on the same issues.

\paragraph{Schedule.}
The default schedule runs all 9 \texttt{tablib} issues first, then all 10 \texttt{tenacity} issues. This blocked structure creates two within-repository learning opportunities with a repository switch in between. A learning agent should build repository-specific knowledge during each block and then reset that knowledge (while retaining general debugging strategies) at the switch.

\paragraph{Dataset and curation.}
Issue instances are drawn from SWE-bench~\citep{jimenez2024swebenchlanguagemodelsresolve}, whose curation pipeline collects merged pull requests from open-source Python repositories and pairs each with a base commit, a reference patch, and a regression test patch. We select two repositories—\texttt{jazzband/tablib} (MIT License) and \texttt{jd/tenacity} (Apache 2.0 License)—whose permissive licenses allow redistribution of task instances. From the full SWE-bench pool for each repository we apply additional filters to produce a tight, learnable sequence: we require a substantive problem statement ($\geq$600 characters), a meaningful code patch (8--300 changed lines), and a non-trivial test patch ($\geq$4 changed lines), and we exclude issues that touch only documentation or tests. From the surviving issues we select the most recent contiguous chronological window that is temporally coherent (total span $\leq$365 days, maximum gap between consecutive issues $\leq$120 days). This windowing ensures that the sequence reflects a realistic period of active development during which repository-level knowledge accumulates naturally.

\paragraph{Example interaction.}\leavevmode
\begin{tcolorbox}[colback=gray!5!white, colframe=gray!50!black,
    fonttitle=\bfseries, fontupper=\small\ttfamily, title=Codebase Adaptation: Issue 4 of 19 (tablib)]
Repository: jazzband/tablib\\
Issue: Dataset.headers setter does not validate that the length of the\\
provided list matches the number of columns in existing rows, allowing\\
silent data corruption.\\[2pt]
Your workspace is at /repo.  Submit your fix with SUBMIT\_PATCH.
\end{tcolorbox}

\begin{tcolorbox}[colback=blue!4!white, colframe=blue!50!black,
    fonttitle=\bfseries, fontupper=\small\ttfamily, title=Codebase Adaptation: Agent session (excerpt)]
\$ grep -n "def headers" tablib/core.py\\
247: @headers.setter\\
\$ sed -n '247,265p' tablib/core.py\\
...\\
\$ python -m pytest tests/test\_core.py::TestDataset::test\_headers -x -q\\
PASSED (1 passed in 0.12s)\\
\$ SUBMIT\_PATCH
\end{tcolorbox}

By issue 4 within the \texttt{tablib} block, a learning agent has already located \texttt{tablib/core.py} as the primary source file, identified \texttt{tests/test\_core.py} as the relevant test module, and knows the project's pytest invocation style---reducing exploratory steps compared to instance 1.

\paragraph{Scoring.}
The per-issue reward is based on step efficiency:
\begin{equation*}
  r = 1 - \frac{\text{steps\_to\_solve} - 1}{\text{max\_steps}}
\end{equation*}
A solved issue earns reward proportional to how early in the step budget the patch was submitted (in this task, max steps was set to 40). An unsolved issue (failed patch, timeout, or exhausted budget) earns reward 0.

\paragraph{Reference maximum.}
$r_{max} = 1.0$, achieved by a system that solves every issue in a single step.

\subsection{Cohort Studies}
\label{appendix:cohort}

\paragraph{Continual learning framing.}
The agent acts as an assistant epidemiologist studying a fictional disease, Vardan-A Syndrome (VAS), with the goal of predicting patient survival at multiple time points from disease diagnosis. The analyst receives data from five independent studies released one at a time. After each study the agent must publish a population-level survival table for 36 predefined patient cohorts, including cohorts the current study cannot observe directly but can have inferred survival curves based on shared traits. Importantly, each study has its own set of demographic variables, coding conventions and biomarkers, but across studies there is a shared set that is predictive of survival curves. In addition, confounding variables can lead to overly confident but incorrect estimates if studies are simply inferred from one at a time without integrating evidence across studies. 

Thus, in this task, continual learning requires a cross-study model of the latent factors that mediate survival. Although the latent factors are stable, there is concept drift from the point of view of the analyst, with conclusions that should change as more studies arrive.

\paragraph{Schedule.}
The default schedule runs 20 instances across five study blocks of four instances each:

\begin{itemize}[noitemsep]
  \item \textbf{HERALD} (4 instances): suburban/rural longitudinal panel; under-represents rapid progressors.
  \item \textbf{MERIDIAN} (4 instances): hospital-based registry; over-represents metabolic comorbidities.
  \item \textbf{MOSAIC} (4 instances): health-fair convenience sample; first study with environmental exposure and metabolic ratio jointly observable.
  \item \textbf{FORGE} (4 instances): workplace-based; heavy on high-exposure groups; drops the cognitive instrument.
  \item \textbf{CADENCE} (4 instances): community screening; over-represents healthy volunteers.
\end{itemize}

Each study measures a partially overlapping subset of the canonical variable set, uses its own column names and coding conventions, and enrolls a biased patient sample. The agent receives no cross-study dictionary: mapping this study's columns onto prior knowledge is part of the task.

\paragraph{Data room and tools.}
Each study instance arrives as one SQLite database with patient-level records (study-specific columns and variable coding, plus a survival outcome), per-column descriptions and coding notes (the agent's primary bridge between cross-study knowledge and local column names), and a brief study description. Agents are also given a consistent set of 5 tools it can use on any given turn: 

\begin{itemize}[noitemsep]
  \item \textbf{get\_database\_metadata}: Variable availability and coding conventions for this study.
  \item \textbf{get\_data\_summary}: Univariate statistics for each column.
  \item \textbf{query\_sql}: Read-only SELECT against the patients table.
  \item \textbf{estimate\_survival\_by\_group}: Per-group survival probabilities at 12, 24 and 36 months (where the group argument is a SQL CASE expression)
  \item \textbf{submit\_cohort\_report}: To submit their final report.
\end{itemize}

\paragraph{Example interaction.}\leavevmode
\begin{tcolorbox}[colback=gray!5!white, colframe=gray!50!black,
    fonttitle=\bfseries, fontupper=\small\ttfamily, title=Cohort Studies: Instance 5 (MERIDIAN Study)]
You are a meta-analyst for the VAS Research Consortium.\\
Study: MERIDIAN Registry | Region: Suburban | Patients: 214\\
Action budget: 20 tool calls.\\[2pt]
Publish updated population survival S(12), S(24), S(36) for\\
all 36 cohorts.  No cross-study references are provided;\\
use your notes from prior studies.
\end{tcolorbox}

\begin{tcolorbox}[colback=blue!4!white, colframe=blue!50!black,
    fonttitle=\bfseries, fontupper=\small\ttfamily, title=Cohort Studies: Agent session (excerpt)]
> get\_database\_metadata()\\
\{columns: \{age\_yrs, metab\_ratio, bp\_sys, serum\_exp, geno3, event, time\}, ...\}\\[2pt]
> query\_sql("SELECT geno3, AVG(event) FROM patients GROUP BY geno3")\\
\{rows: [["low",0.48],["med",0.31],["high",0.19]], ...\}\\[2pt]
> estimate\_survival\_by\_group("CASE WHEN geno3='high' THEN 'R' ELSE ...")\\
\{group\_km: ..., cohort\_split: ...\}\\[2pt]
> submit\_cohort\_report(\{"cohort\_01\_\_s12": 0.79, ..., "cohort\_36\_\_s36": 0.18\})
\end{tcolorbox}

A learning agent cross-references MERIDIAN's \texttt{geno3} column with the three-level genotype encoding it discovered in HERALD and updates its running profile-mixture model accordingly. Cohorts involving genotype are then estimated via the updated cross-study model rather than MERIDIAN's biased in-sample curves.

\paragraph{Scoring.}
The instance score is the information gain (in bits) of the agent's predictions over a naive baseline that reports the current study's overall average survival for every cohort:
\begin{equation*}
  \text{score} = \overline{KL}_{\text{baseline}} - \overline{KL}_{\text{agent}}
\end{equation*}
where $\overline{KL}$ is the mean per-cohort Kullback-Leibler divergence between ground truth and estimated survival probabilities ($KL(P_{\text{true}} \| P_{\text{pred}})$). Averages are across all 36 cohorts and three time points. Positive scores indicate the agent's predictions are closer to population truth than the flat study-wide baseline. The agent receives no score feedback between instances; all learning must come from internal cross-study consistency.

\paragraph{Reference maximum.}
$r_{max} = 0.162$ bits/cohort, estimated by computing the mean per-instance ceiling across the default schedule using oracle predictions derived from the true latent survival model. The ceiling varies by study because each study's sampling bias determines how far the in-study average departs from population truth; this value is the schedule average.

\subsection{Database Exploration}
\label{appendix:dbex}

\paragraph{Continual learning framing.}
The agent answers a sequence of 40 natural-language questions about an unknown SQLite database containing Amazon product reviews across three product groups: Office Products, Electronics, and Musical Instruments. Each question is answered by issuing exploratory SQL queries followed by a final \texttt{ANSWER} action. The schema is not provided upfront---the agent must discover table structure, encoding conventions, and cross-group differences through exploration. The CL challenge is \emph{efficiency}: a learning agent memorizes schema facts (e.g., which group stores prices in cents, which tables lack attribute entries, which timestamp format applies) and reuses them to answer later questions with fewer exploratory queries.

\paragraph{Dataset and Task Sources} Data for this task is sourced from the Amazon Products Metadata and Reviews dataset~\cite{hou2024bridging}, sampling 50K products each from the Office Products, Electronics, and Musical Instruments categories into a single SQLite database of 10 tables. On this data, a set of 40 questions are handwritten by a domain expert, with 20 questions appearing before the database schema drifts and 20 questions appearing after. These questions require knowledge of the database and its quirks to answer with an appropriate SQL query.

\paragraph{Dataset Transformations} To provide an opportunity to learn in a challenging setting, we apply an initial set of data transformations on the raw Amazon products data:

\begin{itemize}
    \item Column names are obfuscated and abbreviated (e.g., \texttt{price -> prc}, \texttt{timestamp -> ts}) and are shared across product categories, masking per-category semantic differences.
    \item Units are modified from the source data (e.g. prices are stored as integer cents in the Electronics category but floating-point dollars in the other two).
    \item Table structure varies across product categories: (e.g. the Electronics category has no attributes table and the Musical Instruments category has no taxonomy table). Queries that may work for one product category may not generalize to other product categories.
    \item Distractor columns are injected to penalize shallow exploration (e.g. a stale 30-day average rating populated for only 30\% of Office Products). 
\end{itemize}

We additionally apply a set of of data migration transformations after question 20, which are discussed in the next section.

\paragraph{Schedule and concept drift.}
The default schedule uses the \texttt{schema\_drift} variant: 20 questions on the original database, then 20 questions after a live migration that introduces a modified database for the agent to deal with, with both schema and value changes. The post-migration questions test whether the agent can notice that previously cached database information no longer applies. An agent that blindly trusts pre-migration facts will continue to issue incorrect queries against renamed or reformatted tables. 

The following set of data transformations are performed to simulate a database migration, which the agent must detect and adapt to after question 20:

\begin{itemize}
    \item A new \texttt{prc\_v2} price column is added to the Electronics products table with updated product prices, while the old price column is kept around.
    \item A new \texttt{verified\_status} text column is partially migrated into the Office Products reviews table, with only 70\% of rows populated.
    \item The Musical Instruments review timestamp column is split into separate year and month integer columns, and 30\% of the original timestamp values are nulled out.
    \item The Musical Instruments attributes table is renamed with a `\_legacy` suffix, with rows dropped from the table as well. This table is no longer meant to be used as a new canonical attributes table is introduced.
\end{itemize}

\paragraph{Example interaction.}\leavevmode
\begin{tcolorbox}[colback=gray!5!white, colframe=gray!50!black,
    fonttitle=\bfseries, fontupper=\small\ttfamily, title=Database Exploration: Question 8 of 40]
Question 8/40\\[2pt]
How many verified purchases are there in the Electronics dataset?\\[2pt]
You may run up to 15 exploratory SQL queries before submitting your\\
final answer.  Queries used so far: 0/15.
\end{tcolorbox}

\begin{tcolorbox}[colback=blue!4!white, colframe=blue!50!black,
    fonttitle=\bfseries, fontupper=\small\ttfamily, title=Database Exploration: Agent response (excerpt)]
action: QUERY\\
content: SELECT COUNT(*) FROM g2\_reviews WHERE vrf = 1\\[4pt]
Query result (1/15 queries used, 7 remaining):\\
\texttt{[(18743,)]}\\[4pt]
action: ANSWER\\
content: 18743
\end{tcolorbox}

A learning agent that already recorded from earlier questions that Electronics is group \texttt{g2}, its reviews table is \texttt{g2\_reviews}, and the verified-purchase column is \texttt{vrf} (integer \texttt{1}/\texttt{0}) needs only a single query to answer this question. A stateless agent would need two or three exploratory queries to rediscover the same facts.

\paragraph{Scoring.}
Per-question reward is based on exploratory query count:
\begin{equation*}
  r = 1 - \frac{\text{exploratory queries used}}{\text{max queries per question}} \quad \text{(correct answer)}
\end{equation*}
An incorrect, timed-out, or over-budget answer incurs the full query budget as regret ($r = 0$). The run score is $1 - \text{cumulative regret} / (N \times \text{max queries})$, where $N = 40$ and the max queries per question is 15. A score of 1.0 requires answering every question correctly with zero exploratory queries; a score of 0.0 means every question was answered incorrectly or exhausted the query budget.

\paragraph{Reference maximum.}
$r_{max} = 1.0$, achieved by a system that answers every question correctly using zero exploratory queries.

\subsection{Exploitable Poker}
\label{appendix:poker}

\paragraph{Continual learning framing.}
The agent plays a sequence of heads-up Texas Hold'em hands against deterministic opponent variants. Each opponent has a fixed, exploitable policy that can be inferred from its betting patterns and showdown reveals. Stacks reset at the start of every hand so that outcomes are comparable across systems and run orders; learning is nonetheless possible because the agent can carry strategy updates forward via its external memory or system prompt. A purely reactive agent that treats each hand independently cannot build an opponent model and leaves a substantial profit edge uncaptured.

Three opponent types are used:
\begin{itemize}[noitemsep]
  \item \textbf{calling\_station}: calls all bets; never bluffs or folds.
  \item \textbf{fit\_or\_fold}: continues only with strong equity; over-folds on marginal hands.
  \item \textbf{loose\_aggressive}: wide opens, frequent continuation bets, sharp tighten on later streets.
\end{itemize}

\paragraph{Game mechanics and hand generation.}
Each hand is a complete game of heads-up Texas Hold'em implemented via the \texttt{texasholdem} Python library. Both players start every hand with a reset 1000-chip stack (small blind 5, big blind 10), so chip outcomes are comparable across all hands and run orders (and baseline scores can be computed for instances in isolation). Cards are dealt by a pseudo-random number generator seeded independently per stage; the seed is fixed, so every hand in the sequence is deterministic and reproducible. The opponent plays a fully deterministic rule-based policy (no randomization), so any variation in outcomes across hands comes entirely from the dealt cards and the agent's actions.

\paragraph{Schedule.}
The default schedule runs 120 hands across five stages: calling station (hands 1--20), loose-aggressive (21--50), return to calling station (51--60), fit-or-fold (61--95), and return to loose-aggressive (96--120). Each stage uses an independent fixed RNG seed so that the dealt hands within a stage are stable across benchmark runs. The return stages test both retention (does the agent recall the earlier opponent model?) and faster adaptation (does re-exposure to a known opponent recover more quickly than the initial exposure?).

\paragraph{Example interaction.}\leavevmode
\begin{tcolorbox}[colback=gray!5!white, colframe=gray!50!black,
    fonttitle=\bfseries, fontupper=\small\ttfamily, title=Exploitable Poker: Hand 3 of 120 (Stage 1: calling station)]
Hand 3/120 | Stacks: You 1000, Opponent 1000 | Blinds: 5/10\\
Your hole cards: A$\spadesuit$ K$\heartsuit$\\
Board (flop): 8$\spadesuit$ 3$\heartsuit$ 2$\spadesuit$\\
Pot: 30.  Action: You to act.  Legal: FOLD, CALL, RAISE.\\[2pt]
Recent history: Hand 1 opponent called pot-size bet on flop and river.\\
Hand 2 opponent called all streets; you won at showdown.
\end{tcolorbox}

\begin{tcolorbox}[colback=blue!4!white, colframe=blue!50!black,
    fonttitle=\bfseries, fontupper=\small\ttfamily, title=Exploitable Poker: Agent response]
\{\\
  "thinking": "Opponent has called large bets in both prior hands\\
    without folding. With top-pair top-kicker and the nut-flush draw\\
    I should bet large to extract value from a station.",\\
  "action": "RAISE",\\
  "amount": 90\\
\}
\end{tcolorbox}

By hand 3 a learning agent has inferred the calling-station pattern from two prior showdowns and adjusts bet sizing upward for value extraction. A stateless agent with no prior context would use a smaller default bet.

\paragraph{Scoring.}
The primary score is average profit in big blinds per hand:
\begin{equation*}
  \text{score} = \frac{1}{N}\sum_{t=1}^{N} \frac{\text{chip profit}_t}{\text{big blind}}
\end{equation*}
Positive scores mean the system won chips on average. The task also records first-half and second-half averages to measure within-run improvement, and per-stage averages to measure adaptation at opponent transitions.

\paragraph{Reference maximum.}
$r_{max} = 9.49$ bb/hand, estimated by simulating an expert policy that knows each opponent's deterministic strategy (but not hidden cards or future runouts) against the same fixed-seed hands in the default schedule.

\subsection{Sales Prediction}
\label{appendix:sales}
\paragraph{Continual learning framing.}
The agent acts as an AI forecasting analyst for a fictional boutique furniture retailer operating three stores in different cities in the United States (San Francisco, New York, Chicago). Each year (instance), the agent receives one store's sales history for the past year only---older exports are discarded by the data warehouse---plus the full product catalog, and must produce a 5-year demand forecast for 75 product-location-year combinations, including stores and products not in the visible data. 

The stable signal to be learned is constant per-cluster growth rates, price-demand relationships, and statistical interactions between product types and locations that define clusters. However, from the agent's perspective the data can look nonstationary: the visible store rotates, the point-of-service schema changes by location, and distractor columns change by stage. The apparent drift is reducible by inferring the persistent latent structure.

The task is implemented in a mini-SWE agent environment with Python libraries installed (numpy, scipy and scikit-learn) for predictive model development. Importantly, the task workspace is carried over across instances so learning agents have the opportunity to persist code and even trained models.

\paragraph{Schedule.}
The default schedule is 12 instances spanning forecast years 2027--2038 in three stages of four instances each. The visible store rotates across all three locations (SF $\to$ NY $\to$ CHI in stage 1, etc.). Each stage also introduces a new set of distractor columns (\texttt{weather\_index}, \texttt{marketing\_spend}, etc.) that are plausible but non-causal. 

\paragraph{Example interaction.}\leavevmode
\begin{tcolorbox}[colback=gray!5!white, colframe=gray!50!black,
    fonttitle=\bfseries, fontupper=\small\ttfamily, title={Sales Prediction: Instance 4 (Year 2030, New York visible)}]
This is year 2030.  Visible store: New York.\\
data/ny\_sales\_2029.csv — columns: store\_id, sku, qty\_sold,\\
  price\_usd, month, marketing\_spend\\
data/furniture.json, data/furniture\_types.json,\\
data/locations.json (New York only)\\[2pt]
Predict 5-year demand for 75 product-location-year combinations\\
including Chicago and San Francisco entries.\\
Score: composite\_score = 1 - |pred - actual| / |actual|.
\end{tcolorbox}

\begin{tcolorbox}[colback=blue!4!white, colframe=blue!50!black,
    fonttitle=\bfseries, fontupper=\small\ttfamily, title=Sales Prediction: Agent session (excerpt)]
\$ python load\_panel.py  \# appends 2029 NY data to persistent panel\\
Panel now covers: SF 2027-2028, CHI 2028-2029, NY 2029\\
\$ python fit\_glm.py     \# Poisson GLM on accumulated panel\\
Cluster growth rates: seating=0.042, tables=0.031, storage=0.018, ...\\
\$ python predict.py     \# 5-year forecasts for all 75 entities\\
\$ echo COMPLETE\_TASK\_AND\_SUBMIT\_FINAL\_OUTPUT
\end{tcolorbox}

\begin{tcolorbox}[colback=blue!4!white, colframe=blue!50!black,
    fonttitle=\bfseries, fontupper=\small\ttfamily, title=Sales Prediction: Structured output (excerpt)]
\{\\
  "predictions": [\\
    \{"locality": "Chicago", "furniture\_name": "Back Bay Accent Chair",\\
     "year": 2031, "items\_sold": 147\},\\
    \{"locality": "San Francisco", "furniture\_name": "Summit Dining Table",\\
     "year": 2032, "items\_sold": 89\},\\
    ...\\
  ]\\
\}
\end{tcolorbox}

By instance 4 a learning agent has accumulated three years of data across two stores, fitted a Poisson GLM that encodes cross-location cluster growth rates, and can produce calibrated out-of-assortment forecasts. A stateless agent refitting from a single store-year produces poorly constrained growth slopes and scores significantly lower.

\paragraph{Scoring.}
Each instance is scored on its full 75-cell submission using Weighted Absolute Percentage Error (WAPE) skill:
\begin{equation*}
  r = 1 - \frac{\sum |\text{predicted} - \text{actual}|}{\sum |\text{actual}|}
\end{equation*}
A perfect forecast earns 1.0; predicting zero everywhere earns 0.0; catastrophic over-prediction yields negative scores. WAPE skill is robust to per-row zeros, invariant to scale, and weights each miss in proportion to its share of total volume. Steps-to-submission is also tracked as a secondary efficiency signal.

\paragraph{Reference maximum.}
$r_{max} = 1.0$, achieved by a system that produces a perfect zero-error forecast for all 75 product-location-year combinations.
\section{Task-Specific Results}
\label{appendix:task_results}

\subsection{Noise Reduction}
\label{sec:noise}

To reduce variance from instance ordering in the benchmark, stateful rollouts are resampled by permuting the instance order within each variant (or by sampling for sequences where the order is exactly specified by the task developer).

Final metrics are averaged across 5 rollouts for all tasks (with the exception of the Codex system due to resource limitations), with 95\% confidence intervals provided for all results. The stateless baseline is evaluated once per task-system pair; since stateless instances are order-independent by construction, no resampling is needed.

\subsection{System Selection Rationale}
\label{sec:sys_selection}

We note that as each full evaluation is run 5 times on 300+ instances across six tasks with each instance having multiple steps; at that scale, exhaustively crossing every available model with every memory system would require thousands of more dollars and runs and is beyond the scope of this initial evaluation. We therefore prioritize coverage of distinct memory paradigms over exhaustive model coverage: full-context ICL is evaluated across our full set of frontier model families to characterize base-model effects, while structured memory systems (Notepad, Mem0, ACE) are paired primarily with GPT-5.4 to enable a controlled within-model comparison. Smaller models and open-weight systems are outside the scope of this paper; we view broader model coverage as a natural direction for future work and expect it to grow as \bench{} is adopted as a standing evaluation track.

\subsection{Normalized Results Per Task}

\begin{figure*}[h]
\vskip -0.1in
\begin{center}
\centerline{\includegraphics[width=0.92\linewidth]{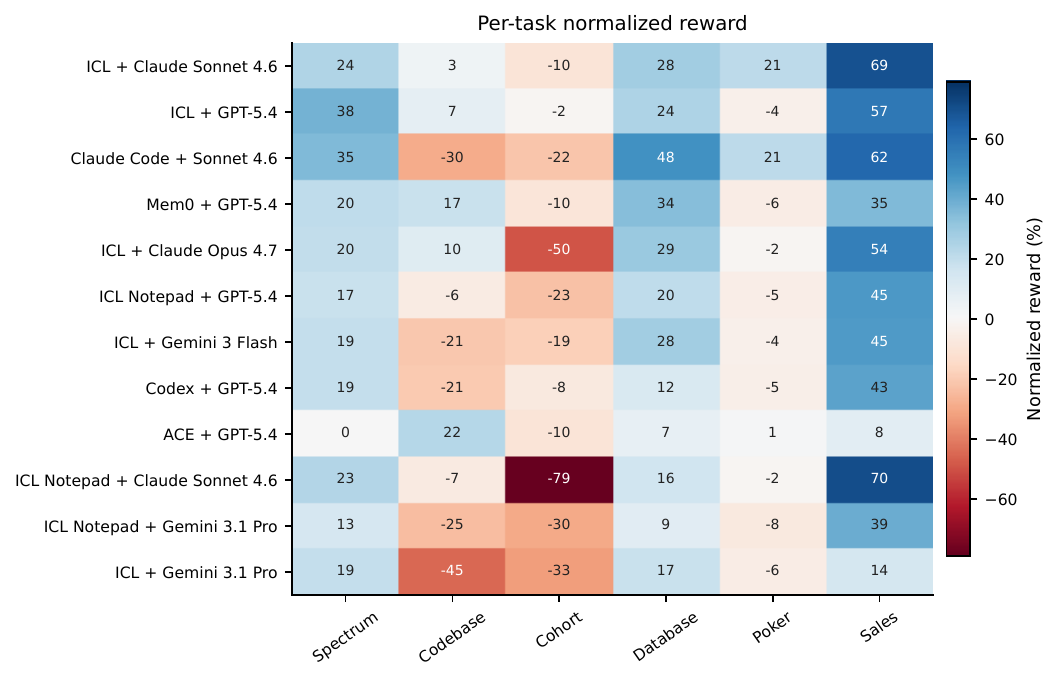}}
\caption{\textbf{Per-task normalized reward across the \bench{} leaderboard.} Values are percentages under the same normalization as \cref{tab:overall-results}; warmer cells indicate stronger relative performance and cooler cells indicate below-baseline performance.}
\label{fig:overall-task-heatmap}
\end{center}
\vskip -0.2in
\end{figure*}

\cref{fig:overall-task-heatmap} shows per-system normalized reward across all six tasks under the same normalization as \cref{tab:overall-results}, providing a compact view of which systems are strongest on which tasks.

\subsection{Task Specific Results}

The tables below report cumulative reward and gain on each task's native reward scale—the raw total rewards accumulated over all instances in the run, without normalization. This makes it straightforward to interpret results for within task comparisons. Bold indicates the best value in each column. All systems were averaged over 5 runs for each task with the exception of the Codex system due to resource constraints. We also provide pareto charts with all systems, and the stateful vs. stateless learning curves for the system that is best on each individual task.

\subsection{Blind Spectrum Monitoring}
\begin{table}[H]
\caption{\textbf{Blind Spectrum Monitoring results.} Cumulative reward and gain are totals across all 90 instances on the task's native reward scale, reported $\pm$ standard error across rollouts. Gain is the cumulative difference between stateful and stateless reward. Cost is the total API cost in USD for a complete task run. Bold indicates the best value in each column.}
\label{tab:task-blind-spectrum-monitoring}
\vskip 0.1in
\begin{center}
\begin{small}
\begin{sc}
\resizebox{\linewidth}{!}{
\begin{tabular}{ll|cccc}
\toprule
System & Model & Cumul. reward & Cumul. gain & Cost (\$) & Runs \\
\midrule
ICL & GPT-5.4 & \textbf{46.198} $\pm$ 1.001 & \textbf{26.437} $\pm$ 1.001 & 1.9 & 5 \\
Claude Code & Sonnet 4.6 & 44.282 $\pm$ 1.449 & 24.522 $\pm$ 1.449 & 10.4 & 5 \\
ICL & Claude Sonnet 4.6 & 36.584 $\pm$ 1.262 & 16.825 $\pm$ 1.262 & 3.6 & 5 \\
ICL Notepad & Claude Sonnet 4.6 & 35.993 $\pm$ 2.414 & 16.233 $\pm$ 2.414 & 3.0 & 5 \\
Mem0 & GPT-5.4 & 33.794 $\pm$ 2.986 & 14.033 $\pm$ 2.986 & 1.4 & 5 \\
ICL & Claude Opus 4.7 & 33.572 $\pm$ 3.082 & 13.813 $\pm$ 3.082 & 7.6 & 5 \\
ICL & Gemini 3 Flash & 33.039 $\pm$ 0.879 & 13.279 $\pm$ 0.879 & \textbf{0.7} & 5 \\
ICL & Gemini 3.1 Pro & 33.033 $\pm$ 1.136 & 13.273 $\pm$ 1.136 & 3.8 & 5 \\
Codex & GPT-5.4 & 32.828 & 13.068 & 3.1 & 1 \\
ICL Notepad & GPT-5.4 & 31.915 $\pm$ 2.122 & 12.153 $\pm$ 2.122 & 1.0 & 5 \\
ICL Notepad & Gemini 3.1 Pro & 29.122 $\pm$ 3.011 & 9.362 $\pm$ 3.011 & 2.8 & 5 \\
ACE & GPT-5.4 & 19.778 $\pm$ 0.009 & 0.017 $\pm$ 0.009 & 4.0 & 5 \\
\bottomrule
\end{tabular}
}
\end{sc}
\end{small}
\end{center}
\vskip -0.1in
\end{table}

\begin{figure}[h]
\begin{center}
\includegraphics[width=\linewidth]{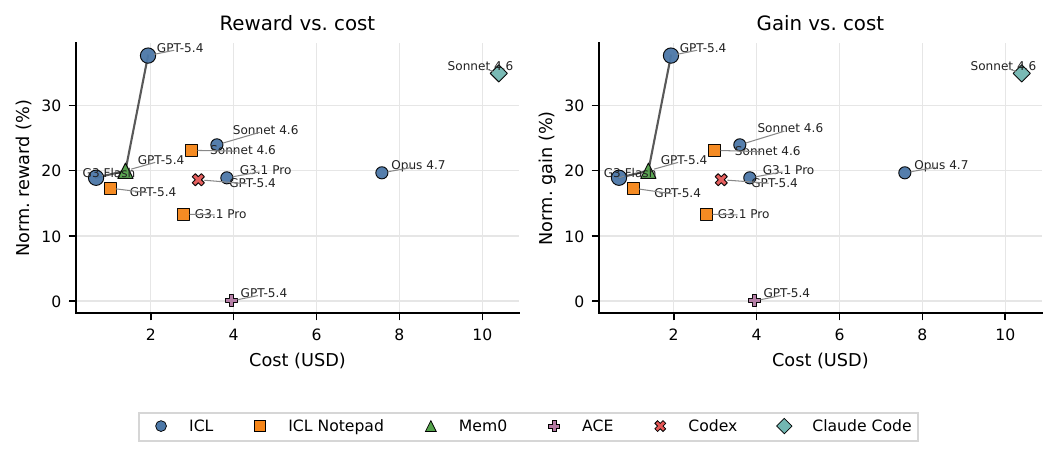}
\end{center}
\caption{\textbf{Blind Spectrum Monitoring: reward and gain vs.\ cost.} Each point is one system configuration; black lines trace the Pareto frontier. Model name abbreviations follow the same convention as \cref{fig:pareto-frontiers}.}
\label{fig:task-pareto-blind-spectrum-monitoring}
\end{figure}

\begin{figure}[h]
\begin{center}
\includegraphics[width=0.8\linewidth]{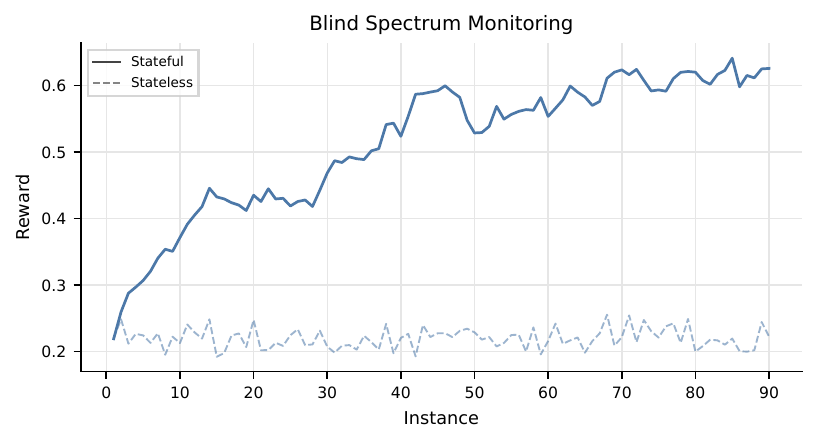}
\end{center}
\caption{\textbf{Blind Spectrum Monitoring: learning curve for the best system on this task (ICL + GPT-5.4).} Solid line shows mean reward per instance in stateful mode; dashed line shows the same system run statelessly on identical instances.}
\label{fig:task-curve-best-blind-spectrum-monitoring}
\end{figure}

\subsection{Codebase Adaptation}
\begin{table}[H]
\caption{\textbf{Codebase Adaptation results.} Cumulative reward and gain are totals across all 19 instances on the task's native reward scale, reported $\pm$ standard error across rollouts. Gain is the cumulative difference between stateful and stateless reward. Cost is the total API cost in USD for a complete task run. Bold indicates the best value in each column.}
\label{tab:task-codebase-adaptation}
\vskip 0.1in
\begin{center}
\begin{small}
\begin{sc}
\resizebox{\linewidth}{!}{
\begin{tabular}{ll|cccc}
\toprule
System & Model & Cumul. reward & Cumul. gain & Cost (\$) & Runs \\
\midrule
ACE & GPT-5.4 & \textbf{11.580} $\pm$ 0.181 & 1.580 $\pm$ 0.181 & 15.7 & 5 \\
Mem0 & GPT-5.4 & 11.105 $\pm$ 1.100 & \textbf{2.980} $\pm$ 1.100 & \textbf{2.6} & 5 \\
ICL & Claude Opus 4.7 & 10.360 $\pm$ 0.924 & 1.485 $\pm$ 0.924 & 8.0 & 5 \\
ICL & GPT-5.4 & 10.140 $\pm$ 1.959 & 0.690 $\pm$ 1.959 & 3.6 & 5 \\
ICL & Claude Sonnet 4.6 & 9.755 $\pm$ 0.223 & 2.705 $\pm$ 0.223 & 6.9 & 5 \\
ICL Notepad & GPT-5.4 & 8.860 $\pm$ 0.685 & 0.610 $\pm$ 0.685 & 3.5 & 5 \\
ICL Notepad & Claude Sonnet 4.6 & 8.765 $\pm$ 0.455 & 0.890 $\pm$ 0.455 & 4.2 & 5 \\
Codex & GPT-5.4 & 7.450 & -0.800 & 3.8 & 1 \\
ICL & Gemini 3 Flash & 7.420 $\pm$ 0.934 & -0.330 $\pm$ 0.934 & 2.8 & 5 \\
ICL Notepad & Gemini 3.1 Pro & 7.110 $\pm$ 0.634 & -1.315 $\pm$ 0.634 & 3.5 & 5 \\
Claude Code & Sonnet 4.6 & 6.630 $\pm$ 1.755 & 0.730 $\pm$ 1.755 & 6.8 & 5 \\
ICL & Gemini 3.1 Pro & 5.125 $\pm$ 0.652 & -1.550 $\pm$ 0.652 & 3.4 & 5 \\
\bottomrule
\end{tabular}
}
\end{sc}
\end{small}
\end{center}
\vskip -0.1in
\end{table}

\begin{figure}[h]
\begin{center}
\includegraphics[width=\linewidth]{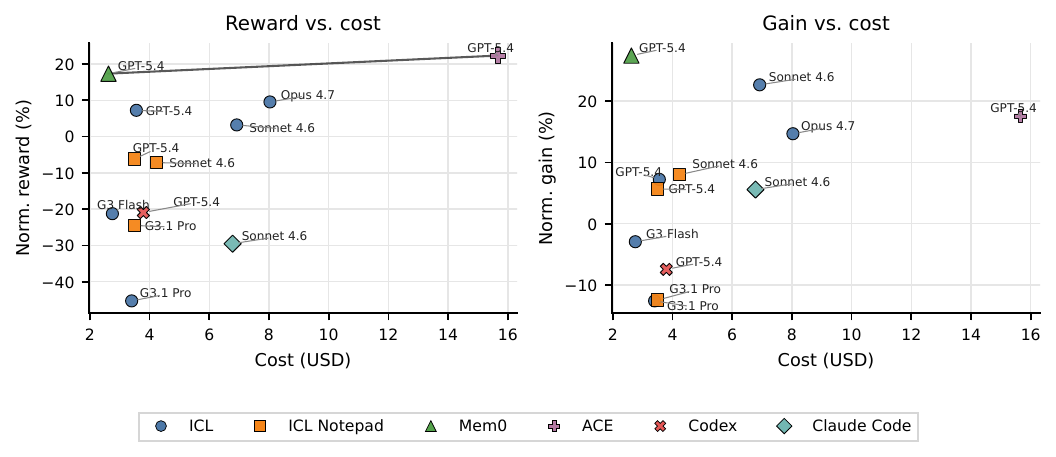}
\end{center}
\caption{\textbf{Codebase Adaptation: reward and gain vs.\ cost.} Each point is one system configuration; black lines trace the Pareto frontier. Model name abbreviations follow the same convention as \cref{fig:pareto-frontiers}.}
\label{fig:task-pareto-codebase-adaptation}
\end{figure}

\begin{figure}[h]
\begin{center}
\includegraphics[width=0.78\linewidth]{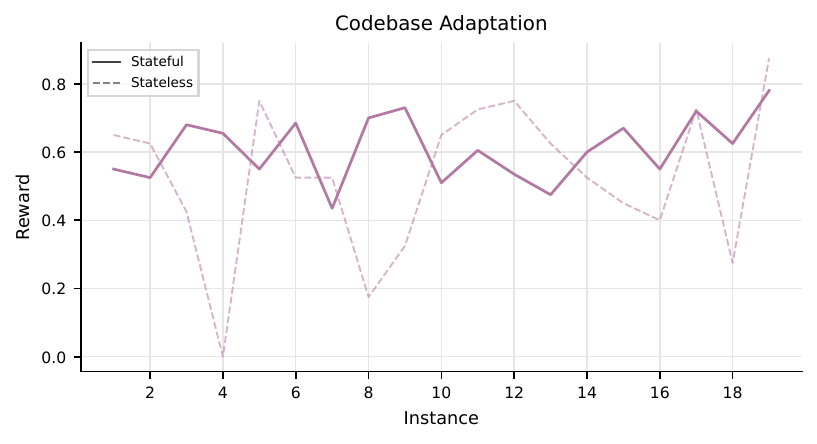}
\end{center}
\caption{\textbf{Codebase Adaptation: learning curve for the best system on this task (ACE + GPT-5.4).} Solid line shows mean reward per instance in stateful mode; dashed line shows the same system run statelessly on identical instances.}
\label{fig:task-curve-best-codebase-adaptation}
\end{figure}

\subsection{Cohort Studies}
\begin{table}[H]
\caption{\textbf{Cohort Studies results.} Cumulative reward and gain are totals across all 20 instances on the task's native reward scale, reported $\pm$ standard error across rollouts. Gain is the cumulative difference between stateful and stateless reward. Cost is the total API cost in USD for a complete task run. Bold indicates the best value in each column.}
\label{tab:task-cohort-studies}
\vskip 0.1in
\begin{center}
\begin{small}
\begin{sc}
\resizebox{\linewidth}{!}{
\begin{tabular}{ll|cccc}
\toprule
System & Model & Cumul. reward & Cumul. gain & Cost (\$) & Runs \\
\midrule
ICL & GPT-5.4 & \textbf{0.957} $\pm$ 0.393 & -0.037 $\pm$ 0.393 & 3.7 & 5 \\
Codex & GPT-5.4 & 0.821 & 0.316 & 7.8 & 1 \\
Mem0 & GPT-5.4 & 0.778 $\pm$ 0.315 & -0.067 $\pm$ 0.315 & 6.0 & 5 \\
ICL & Claude Sonnet 4.6 & 0.762 $\pm$ 0.240 & 0.185 $\pm$ 0.240 & 5.6 & 5 \\
ACE & GPT-5.4 & 0.759 $\pm$ 0.127 & 0.383 $\pm$ 0.127 & 12.8 & 5 \\
ICL & Gemini 3 Flash & 0.576 $\pm$ 0.215 & 0.437 $\pm$ 0.215 & \textbf{1.2} & 5 \\
Claude Code & Sonnet 4.6 & 0.496 $\pm$ 0.226 & -0.309 $\pm$ 0.226 & 7.2 & 5 \\
ICL Notepad & GPT-5.4 & 0.476 $\pm$ 0.046 & -0.900 $\pm$ 0.046 & 4.4 & 5 \\
ICL Notepad & Gemini 3.1 Pro & 0.327 $\pm$ 0.086 & -0.366 $\pm$ 0.086 & 1.5 & 5 \\
ICL & Gemini 3.1 Pro & 0.257 $\pm$ 0.143 & -0.563 $\pm$ 0.143 & 1.8 & 5 \\
ICL & Claude Opus 4.7 & -0.121 $\pm$ 0.093 & -0.091 $\pm$ 0.093 & 7.0 & 5 \\
ICL Notepad & Claude Sonnet 4.6 & -0.784 $\pm$ 0.358 & \textbf{0.795} $\pm$ 0.358 & 11.6 & 5 \\
\bottomrule
\end{tabular}
}
\end{sc}
\end{small}
\end{center}
\vskip -0.1in
\end{table}

\begin{figure}[h]
\begin{center}
\includegraphics[width=\linewidth]{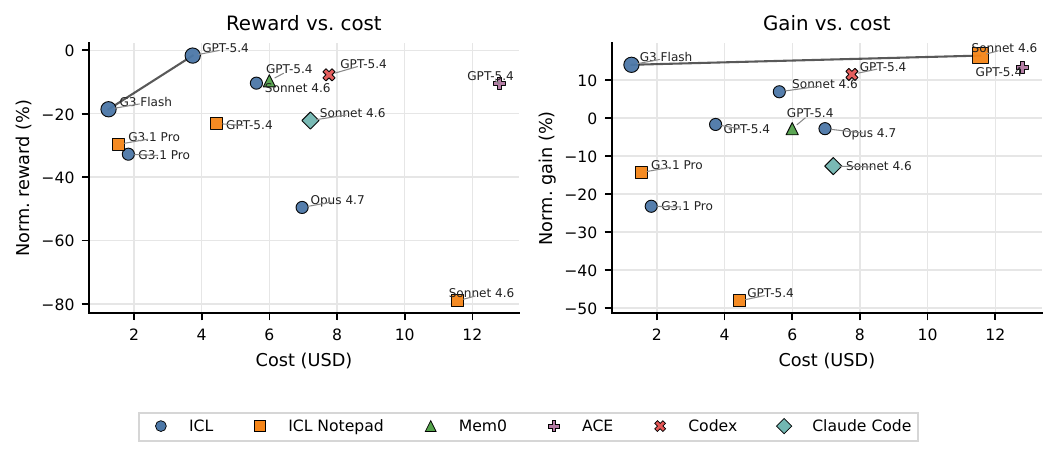}
\end{center}
\caption{\textbf{Cohort Studies: reward and gain vs.\ cost.} Each point is one system configuration; black lines trace the Pareto frontier. Model name abbreviations follow the same convention as \cref{fig:pareto-frontiers}.}
\label{fig:task-pareto-cohort-studies}
\end{figure}

\begin{figure}[h]
\begin{center}
\includegraphics[width=0.78\linewidth]{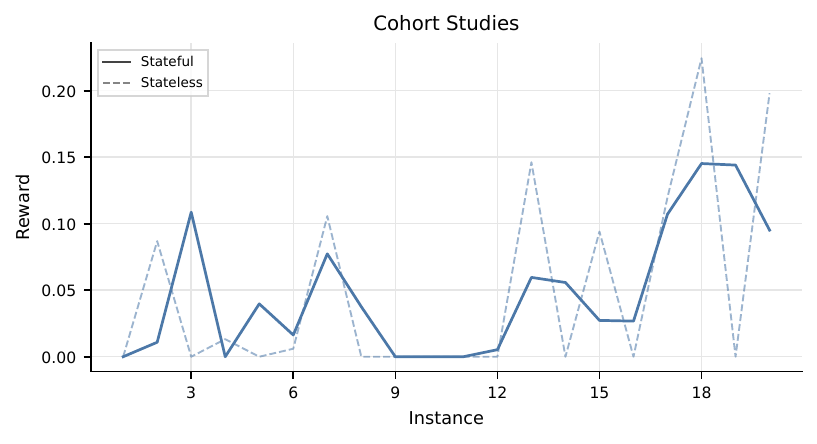}
\end{center}
\caption{\textbf{Cohort Studies: learning curve for the best system on this task (ICL + GPT-5.4).} Solid line shows mean reward per instance in stateful mode; dashed line shows the same system run statelessly on identical instances.}
\label{fig:task-curve-best-cohort-studies}
\end{figure}

\subsection{Database Exploration}
\begin{table}[H]
\caption{\textbf{Database Exploration results.} Cumulative reward and gain are totals across all 40 instances on the task's native reward scale, reported $\pm$ standard error across rollouts. Gain is the cumulative difference between stateful and stateless reward. Cost is the total API cost in USD for a complete task run. Bold indicates the best value in each column.}
\label{tab:task-database-exploration}
\vskip 0.1in
\begin{center}
\begin{small}
\begin{sc}
\resizebox{\linewidth}{!}{
\begin{tabular}{ll|cccc}
\toprule
System & Model & Cumul. reward & Cumul. gain & Cost (\$) & Runs \\
\midrule
Claude Code & Sonnet 4.6 & \textbf{22.053} $\pm$ 1.182 & \textbf{13.853} $\pm$ 1.182 & 3.3 & 5 \\
Mem0 & GPT-5.4 & 17.240 $\pm$ 1.300 & 12.907 $\pm$ 1.300 & 2.0 & 5 \\
ICL & Claude Opus 4.7 & 15.653 $\pm$ 1.781 & 9.587 $\pm$ 1.781 & 5.2 & 5 \\
ICL & Gemini 3 Flash & 15.027 $\pm$ 0.712 & 11.493 $\pm$ 0.712 & \textbf{0.4} & 5 \\
ICL & Claude Sonnet 4.6 & 15.013 $\pm$ 1.427 & 8.480 $\pm$ 1.427 & 1.9 & 5 \\
ICL & GPT-5.4 & 13.880 $\pm$ 1.715 & 8.347 $\pm$ 1.715 & 1.0 & 5 \\
ICL Notepad & GPT-5.4 & 12.373 $\pm$ 1.158 & 6.373 $\pm$ 1.158 & 1.5 & 5 \\
ICL & Gemini 3.1 Pro & 11.560 $\pm$ 2.894 & 6.827 $\pm$ 2.894 & 1.3 & 5 \\
ICL Notepad & Claude Sonnet 4.6 & 11.000 $\pm$ 0.877 & 3.667 $\pm$ 0.877 & 2.2 & 5 \\
Codex & GPT-5.4 & 9.600 & 6.133 & 1.8 & 1 \\
ICL Notepad & Gemini 3.1 Pro & 8.520 $\pm$ 0.651 & 4.320 $\pm$ 0.651 & 2.6 & 5 \\
ACE & GPT-5.4 & 7.853 $\pm$ 1.291 & 2.387 $\pm$ 1.291 & 8.8 & 5 \\
\bottomrule
\end{tabular}
}
\end{sc}
\end{small}
\end{center}
\vskip -0.1in
\end{table}

\begin{figure}[h]
\begin{center}
\includegraphics[width=\linewidth]{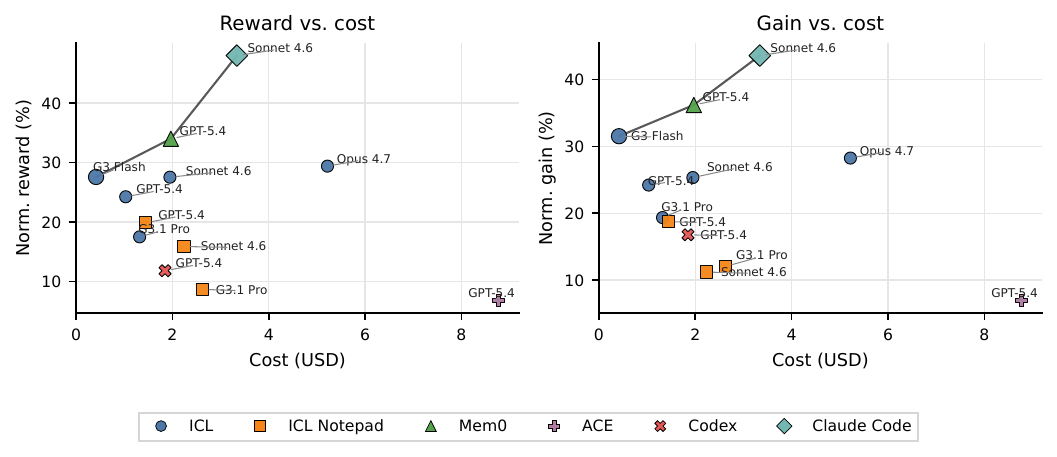}
\end{center}
\caption{\textbf{Database Exploration: reward and gain vs.\ cost.} Each point is one system configuration; black lines trace the Pareto frontier. Model name abbreviations follow the same convention as \cref{fig:pareto-frontiers}.}
\label{fig:task-pareto-database-exploration}
\end{figure}

\begin{figure}[h]
\begin{center}
\includegraphics[width=0.78\linewidth]{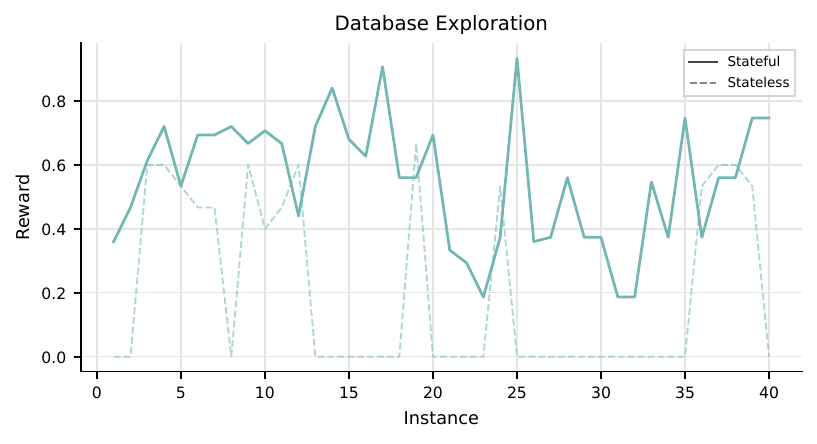}
\end{center}
\caption{\textbf{Database Exploration: learning curve for the best system on this task (Claude Code + Sonnet 4.6).} Solid line shows mean reward per instance in stateful mode; dashed line shows the same system run statelessly on identical instances.}
\label{fig:task-curve-best-database-exploration}
\end{figure}

\subsection{Exploitable Poker}
\begin{table}[H]
\caption{\textbf{Exploitable Poker results.} Cumulative reward and gain are totals across all 120 instances on the task's native reward scale, reported $\pm$ standard error across rollouts. Gain is the cumulative difference between stateful and stateless reward. Cost is the total API cost in USD for a complete task run. Bold indicates the best value in each column.}
\label{tab:task-exploitable-poker}
\vskip 0.1in
\begin{center}
\begin{small}
\begin{sc}
\resizebox{\linewidth}{!}{
\begin{tabular}{ll|cccc}
\toprule
System & Model & Cumul. reward & Cumul. gain & Cost (\$) & Runs \\
\midrule
Claude Code & Sonnet 4.6 & \textbf{343.020} $\pm$ 29.367 & \textbf{58.520} $\pm$ 29.367 & 8.7 & 5 \\
ICL & Claude Sonnet 4.6 & 339.920 $\pm$ 19.530 & 23.220 $\pm$ 19.530 & 9.4 & 5 \\
ACE & GPT-5.4 & 143.500 $\pm$ 41.196 & 1.600 $\pm$ 41.196 & 13.2 & 5 \\
ICL & Claude Opus 4.7 & 116.680 $\pm$ 26.485 & -41.020 $\pm$ 26.485 & 17.5 & 5 \\
ICL Notepad & Claude Sonnet 4.6 & 115.420 $\pm$ 29.931 & -201.580 $\pm$ 29.931 & 6.7 & 5 \\
ICL & GPT-5.4 & 95.760 $\pm$ 26.414 & -37.840 $\pm$ 26.414 & 4.6 & 5 \\
ICL & Gemini 3 Flash & 94.840 $\pm$ 16.912 & -101.960 $\pm$ 16.912 & 1.9 & 5 \\
Codex & GPT-5.4 & 85.000 & 20.500 & 8.3 & 1 \\
ICL Notepad & GPT-5.4 & 81.340 $\pm$ 15.206 & 0.240 $\pm$ 15.206 & 1.5 & 5 \\
ICL & Gemini 3.1 Pro & 76.400 $\pm$ 3.868 & 32.900 $\pm$ 3.868 & 4.0 & 5 \\
Mem0 & GPT-5.4 & 73.420 $\pm$ 41.548 & -16.980 $\pm$ 41.548 & 3.6 & 5 \\
ICL Notepad & Gemini 3.1 Pro & 53.500 $\pm$ 6.684 & 17.000 $\pm$ 6.684 & \textbf{1.3} & 5 \\
\bottomrule
\end{tabular}
}
\end{sc}
\end{small}
\end{center}
\vskip -0.1in
\end{table}

\begin{figure}[h]
\begin{center}
\includegraphics[width=\linewidth]{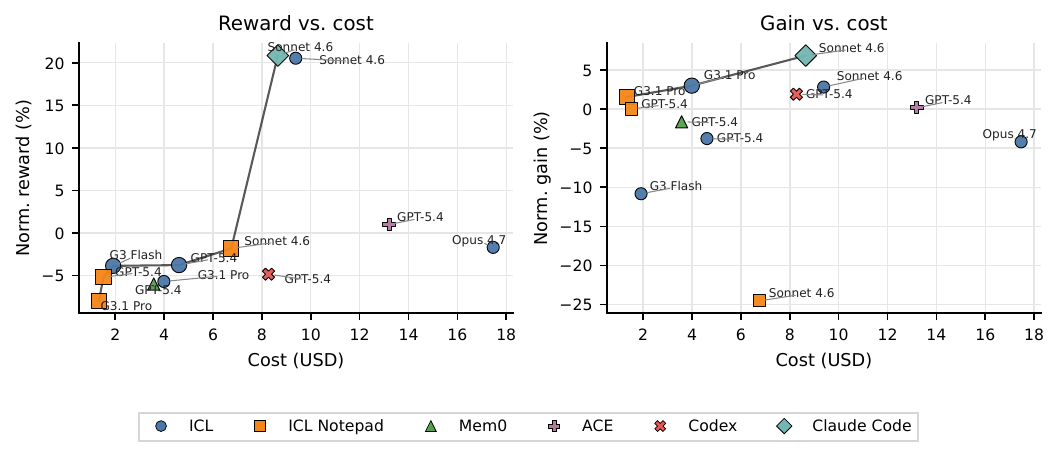}
\end{center}
\caption{\textbf{Exploitable Poker: reward and gain vs.\ cost.} Each point is one system configuration; black lines trace the Pareto frontier. Model name abbreviations follow the same convention as \cref{fig:pareto-frontiers}.}
\label{fig:task-pareto-exploitable-poker}
\end{figure}

\begin{figure}[h]
\begin{center}
\includegraphics[width=0.78\linewidth]{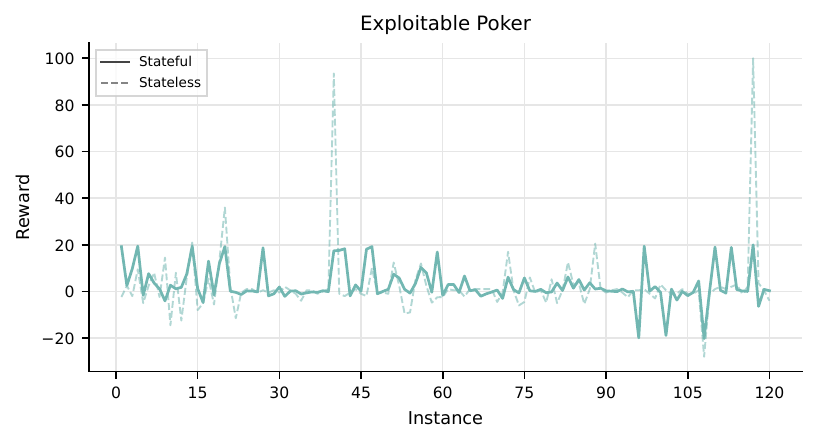}
\end{center}
\caption{\textbf{Exploitable Poker: learning curve for the best system on this task (Claude Code + Sonnet 4.6).} Solid line shows mean reward per instance in stateful mode; dashed line shows the same system run statelessly on identical instances.}
\label{fig:task-curve-best-exploitable-poker}
\end{figure}

\subsection{Sales Prediction}
\begin{table}[H]
\caption{\textbf{Sales Prediction results.} Cumulative reward and gain are totals across all 12 instances on the task's native reward scale, reported $\pm$ standard error across rollouts. Gain is the cumulative difference between stateful and stateless reward. Cost is the total API cost in USD for a complete task run. Bold indicates the best value in each column.}
\label{tab:task-sales-prediction}
\vskip 0.1in
\begin{center}
\begin{small}
\begin{sc}
\resizebox{\linewidth}{!}{
\begin{tabular}{ll|cccc}
\toprule
System & Model & Cumul. reward & Cumul. gain & Cost (\$) & Runs \\
\midrule
ICL Notepad & Claude Sonnet 4.6 & \textbf{10.073} $\pm$ 0.095 & \textbf{5.790} $\pm$ 0.095 & 3.8 & 5 \\
ICL & Claude Sonnet 4.6 & 10.018 $\pm$ 0.215 & 4.840 $\pm$ 0.215 & 3.0 & 5 \\
Claude Code & Sonnet 4.6 & 9.573 $\pm$ 0.254 & 4.537 $\pm$ 0.254 & 2.2 & 5 \\
ICL & GPT-5.4 & 9.230 $\pm$ 0.254 & 3.633 $\pm$ 0.254 & 3.5 & 5 \\
ICL & Claude Opus 4.7 & 9.039 $\pm$ 0.292 & 4.649 $\pm$ 0.292 & 4.4 & 5 \\
ICL Notepad & GPT-5.4 & 8.486 $\pm$ 0.317 & 4.003 $\pm$ 0.317 & 2.4 & 5 \\
ICL & Gemini 3 Flash & 8.481 $\pm$ 0.211 & 3.210 $\pm$ 0.211 & \textbf{0.6} & 5 \\
Codex & GPT-5.4 & 8.320 & 3.144 & 2.4 & 5 \\
ICL Notepad & Gemini 3.1 Pro & 8.108 $\pm$ 0.350 & 4.952 $\pm$ 0.350 & 1.5 & 5 \\
Mem0 & GPT-5.4 & 7.815 $\pm$ 0.198 & 3.064 $\pm$ 0.198 & 2.8 & 5 \\
ICL & Gemini 3.1 Pro & 6.468 $\pm$ 0.054 & 2.560 $\pm$ 0.054 & 0.8 & 5 \\
ACE & GPT-5.4 & 6.116 $\pm$ 0.129 & 0.913 $\pm$ 0.129 & 8.4 & 5 \\
\bottomrule
\end{tabular}
}
\end{sc}
\end{small}
\end{center}
\vskip -0.1in
\end{table}

\begin{figure}[H]
\begin{center}
\includegraphics[width=\linewidth]{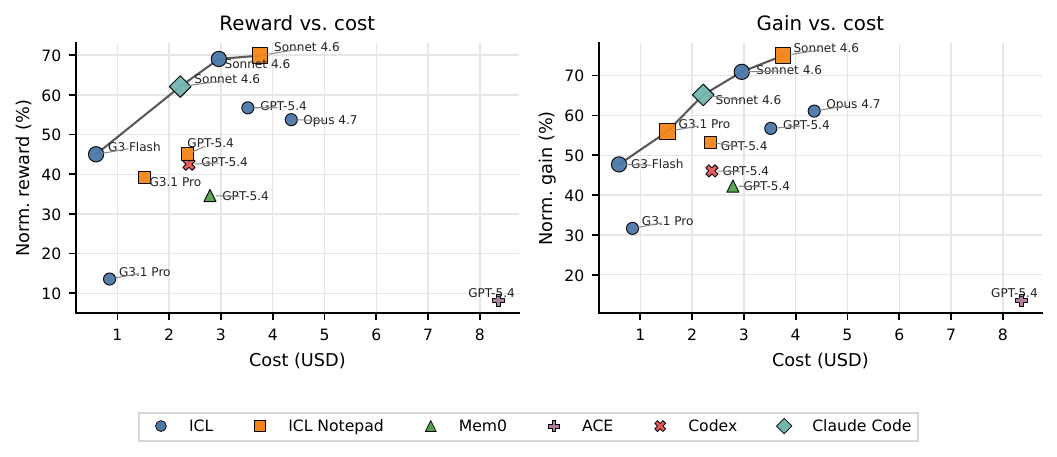}
\end{center}
\caption{\textbf{Sales Prediction: reward and gain vs.\ cost.} Each point is one system configuration; black lines trace the Pareto frontier. Model name abbreviations follow the same convention as \cref{fig:pareto-frontiers}.}
\label{fig:task-pareto-sales-prediction}
\end{figure}

\begin{figure}[H]
\begin{center}
\includegraphics[width=0.78\linewidth]{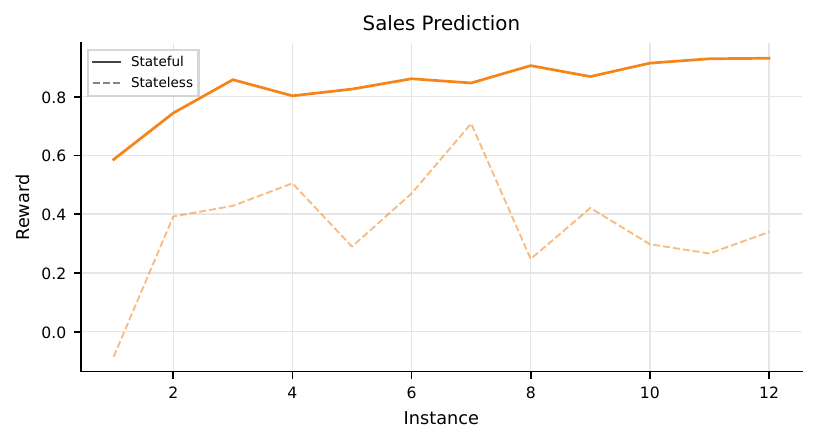}
\end{center}
\caption{\textbf{Sales Prediction: learning curve for the best system on this task (ICL Notepad + Claude Sonnet 4.6).} Solid line shows mean reward per instance in stateful mode; dashed line shows the same system run statelessly on identical instances.}
\label{fig:task-curve-best-sales-prediction}
\end{figure}

\section{Stability--Plasticity Decomposition of Gain}
\label{appendix:learning_decomposition}

The normalized gain $\widehat{g}$ in \cref{sec:cross_task_norm} captures how much a system learns over a run. We decomposed the measure into two terms related to the stability-plasticity tradeoff in continual learning \citep{van_de_Ven_2025, delange2023stabilitygap, delange2022survey}: a "plasticity" term indicating how well agents adapt to new task information with the given memory system, and a "stability" term indicating how well agents re-use older information to benefit new tasks. Two systems can demonstrate the same $\widehat{g}$ while differing with respect to these terms: a highly plastic learner can adapt well within each task variant while failing to carry that adaptation to new variants, while a highly stable learner can retain useful structure across variants but adapt more slowly overall. To separate these effects we decomposed $\widehat{g}$ into two additive shares attributable to within-variant adaptation and across-variant retention respectively.

\subsection{Variant-based Partition}
\label{sec:within_variant_partition}

Each \bench{} task is a sequence of one or more variants (see \ref{sec:cl_bench}), contiguous runs of instances sharing a common set of properties or statistics. Different variants for the same task can differ with respect to these properties or statistics, leading to a non-stationary stream of observations, but importantly, the variants share a common latent structure that is predictive of future data streams. Let $i_t \in \{0, 1, 2, \dots\}$ denote the within-variant index of instance $t$: its position inside the variant that contains it. We partition the $N$ instances of a run into

$$ \mathcal{B} = \{t : i_t = 0\}, \qquad \mathcal{W} = \{t : i_t > 0\}, $$

the boundary (first-instance-of-variant) and within-variant index sets, with fractions $f_\mathcal{B} = |\mathcal{B}|/N$ and $f_\mathcal{W} = 1 - f_\mathcal{B}$ for scaling terms by how much each type is represented over the entire task schedule. Boundary instances are the moments at which the system encounters a new variant and has only its prior accumulated state to draw on; within-variant instances follow at least one round of in-variant feedback.

\subsection{Decomposition}
\label{sec:decomposition_definition}

Restricting the per-instance rewards $r^{sf}_t$ and $r^{sl}_t$ from \cref{sec:reward} to a partition $\mathcal{S} \in \{\mathcal{B}, \mathcal{W}\}$, let $\bar{r}^{sf}_\mathcal{S}$ and $\bar{r}^{sl}_\mathcal{S}$ denote the corresponding stateful and stateless mean rewards. Using the same task-level headroom denominator $r_{max} - \bar{r}^{sl}$ as in $\widehat{g}$ (\cref{sec:cross_task_norm}), we define the stability component:

$$ \widehat{g}_{\text{stab}} \;=\; f_\mathcal{B} \cdot \frac{\bar{r}^{sf}_\mathcal{B} - \bar{r}^{sl}_\mathcal{B}}{r_{max} - \bar{r}^{sl}} $$

and the plasticity share component:

$$ \widehat{g}_{\text{plas}} \;=\; f_\mathcal{W} \cdot \frac{\bar{r}^{sf}_\mathcal{W} - \bar{r}^{sl}_\mathcal{W}}{r_{max} - \bar{r}^{sl}}. $$

It is straightforward to show that:

$$ \widehat{g} \;=\; \widehat{g}_{\text{stab}} \;+\; \widehat{g}_{\text{plas}}. $$

The signs of each term can be interpreted in the same way as $\widehat{g}$ from \cref{sec:gain}. For example, a system that retains useful structure across variant boundaries will indicate $\widehat{g}_{\text{stab}} > 0$ while one that misapplies learned information can yield $\widehat{g}_{\text{stab}} < 0$ even when within-variant adaptation recovers performance.

To derive estimates of these components for pairings of memory systems and agents, we calculated the terms for each task first and then averaged across tasks (so tasks with more instances would not dominate more).

\begin{figure*}[t]
\vskip -0.1in
\begin{center}
\centerline{\includegraphics[width=0.88\linewidth]{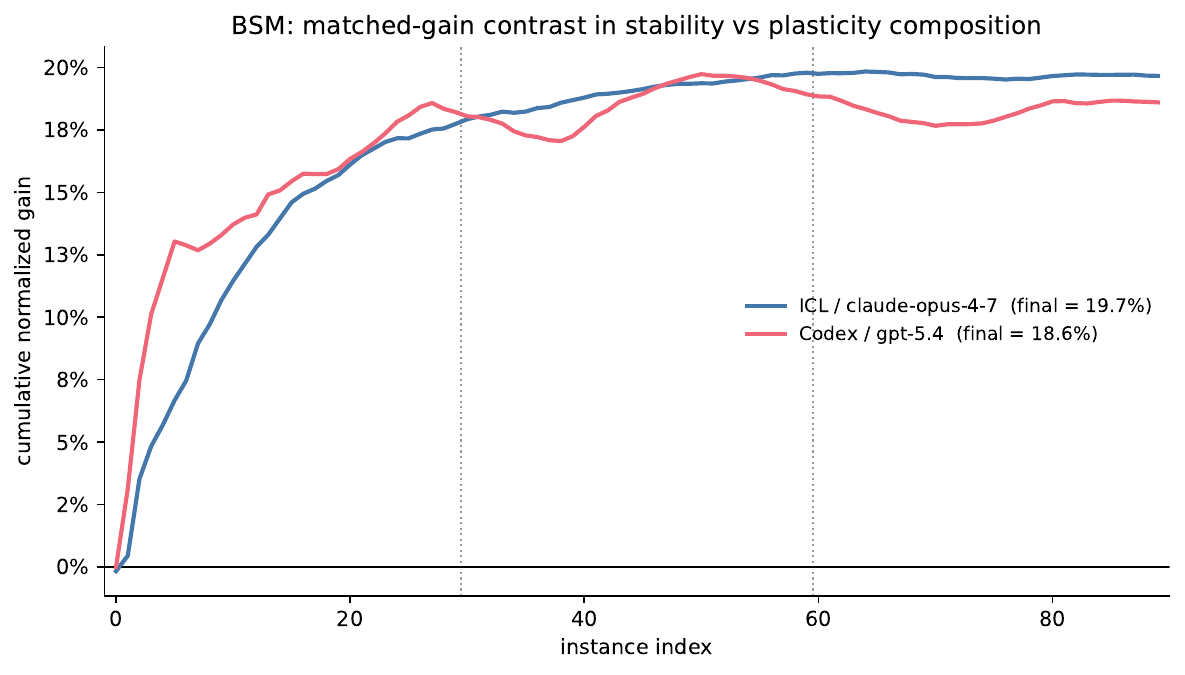}}
\caption{\textbf{Example of two learners with differing amounts of learning stability.} Normalized gain over the blind spectrum monitoring task instances for two systems, one with 13.5\% stability (Claude Opus 4.7 with ICL) and one with significantly lower stability at 9\% of learning headroom (GPT-5.4 with Codex).}
\label{fig:learning-decomposition-example}
\end{center}
\vskip -0.2in
\end{figure*}

Figure \ref{fig:learning-decomposition-example} shows an example of two agents with differing levels of learning stability but similar overall gain.
\section{Agent failure examples}
\label{appendix:failure_examples}

\paragraph{Stability failure example.} In the Sales Prediction task, the agent needs to predict a 5 year forecast each round and is given a composite score between rounds. Going into Round 1, the feedback score from the prior round is 0.945, and the agent decides the previous round overpredicted New York (specifically accent chairs, brass end tables, and king bed frames). It flattens the King Bed Frame series for the next five years. The flat forecast scores 0.936, and the agent now decides it underpredicted that same series. In this example, immediate observations dominated over the agent's prior view of long-run New York demand.

\begin{tcolorbox}[colback=blue!4!white, colframe=blue!50!black,
    fonttitle=\bfseries, fontupper=\small\ttfamily,
    title={Stability failure --- Sales Prediction (Round 1, year 2035)}]
\textit{Environment feedback after the previous round (step 43):}\\[2pt]
``Predictions recorded for year 2034. Composite score: 0.945.\\
\quad Moving to instance 9/12...''\\[6pt]
\textit{Round-1 agent reasoning (step 44):}\\[2pt]
``Because last round overpredicted New York accent/brass and king\\
bed frames\ldots\ I should\ldots\ recalibrate 2035--2039 forecasts.''\\[6pt]
\textit{Round-1 forecast values written by the agent (step 45;\\submitted unchanged at step 47):}\\[2pt]
('New York','Back Bay King Bed Frame'):\\
\quad\{2035:230, 2036:228, 2037:227, 2038:227, 2039:228\}\\[2pt]
$\rightarrow$ Forecast is essentially flat across the next 5 years.
\end{tcolorbox}

\begin{tcolorbox}[colback=blue!4!white, colframe=blue!50!black,
    fonttitle=\bfseries, fontupper=\small\ttfamily,
    title={Stability failure --- Sales Prediction (Round 2, year 2036)}]
\textit{Environment feedback after Round 1 (step 47):}\\[2pt]
``Predictions recorded for year 2035. Composite score: 0.936.\\
\quad Moving to instance 10/12...''\\[6pt]
\textit{Round-2 agent reasoning (step 48):}\\[2pt]
``\ldots Last round underpredicted NY accent/brass and king bed\\
frame, overpredicted Chicago bookcase/desk chair\ldots''\\[6pt]
\textit{Round-2 forecast values for the same series (step 49;\\submitted unchanged at step 51):}\\[2pt]
('New York','Back Bay King Bed Frame'):\\
\quad\{2036:321, 2037:371, 2038:430, 2039:499, 2040:580\}\\[2pt]
$\rightarrow$ Forecast now grows roughly $+16\%$/year for the same product.
\end{tcolorbox}

\paragraph{Plasticity failure example.} In the Cohort Studies task, the agent walks through 20 clinical studies in order, estimating survival for 36 cohorts each time. Each study only measures a subset of the variables, so cohorts whose defining variables are absent in the current study can only be estimated by reusing evidence from earlier studies. When the agent reaches Study~15 (FORGE Urban), it has already seen the full database schemas of HERALD, MERIDIAN, and MOSAIC. 

At Study~15, the agent recognizes that the study is structurally similar to a prior FORGE instance and reuses the within-type model. However, for the 28 cohorts whose defining variables are absent in this dataset, the agent dismisses them as "cohort definitions from a different study schema that don't apply here" and submits the same triplet for every one of them. \emph{The cross-study evidence
the agent observed earlier is not applied to the new instance.} The notepad has relevant content, but the agent does not bring that content forward as context for the right instances.

\begin{tcolorbox}[colback=green!4!white, colframe=green!50!black,
    fonttitle=\bfseries, fontupper=\small\ttfamily,
    title={What the agent observed in earlier studies (the ``notepad'' content)}]
\textit{From Study 1 (HERALD) database schema, step 1:}\\[2pt]
``family\_history\_count: First-degree relatives with neurological\\
\quad disease  units=count''\\
``mmse\_score: Mini-Mental State Examination (MMSE)''\\[6pt]
\textit{From Study 5 (MERIDIAN) database schema, step 59:}\\[2pt]
``prb1\_ratio: PRB1 protein ratio''\\
``moca\_score: Montreal Cognitive Assessment (MoCA)''\\[6pt]
\textit{From Study 9 (MOSAIC) database schema, step 112:}\\[2pt]
``prb1\_category: PRB1 ratio category''\\
``years\_of\_education: Formal education completed''\\[6pt]
\textit{Agent at Study 3 (step 34) referring explicitly to its notepad:}\\[2pt]
``\ldots From my notepad, I have detailed survival data from Study 2\\
which was also a Rural region\ldots''
\end{tcolorbox}

\begin{tcolorbox}[colback=blue!4!white, colframe=blue!50!black,
    fonttitle=\bfseries, fontupper=\small\ttfamily,
    title={Plasticity failure --- Study 15 (FORGE Urban) setup and agent session}]
\textit{Study 15 setup shown to the agent (step 194):}\\[2pt]
``\#\# Study 15/20: FORGE Study\ldots\\
**Region(s) in this dataset:** Urban\\
**Patients:** 148\\
\ldots submit survival estimates (P(survival) at 12, 24, and 36\\
months) for 36 population cohorts.''\\[6pt]
\textit{Agent's opening thought (step 194):}\\[2pt]
``This is Study 15, which is another FORGE Study but Urban only with\\
148 patients - same as Study 13. Let me\ldots\ leverage my knowledge\\
from Study 13 to efficiently analyze this dataset.''\\[6pt]
\textit{After querying Study 15's data (step 197):}\\[2pt]
``The unobservable cohorts reference variables like family\_history,\\
prb1, cognitive, education, avoc\_high, gg3 - which are NOT in this\\
dataset. These seem to be cohort definitions from a different study\\
schema that don't apply here\ldots''\\[6pt]
\textit{Final plan before submission (step 203):}\\[2pt]
``Now I have the overall study survival: S(12)=0.4928, S(24)=0.3171,\\
S(36)=0.2255. This will be used for all 28 unobservable cohorts.''\\[2pt]
$\rightarrow$ Instance reward: 0.005 (essentially the flat-baseline reward).
\end{tcolorbox}


\end{document}